\documentclass[pdflatex,sn-mathphys-num, twocolumn]{sn-jnl}
\usepackage{my_style}
\begin{document}

\title[Article Title]{Automatic Robot Hand-Eye Calibration Enabled by Learning-Based 3D Vision}
\author[1]{\fnm{Leihui} \sur{Li}}\email{leihui@mpe.au.dk}

\author[1]{\fnm{Xingyu} \sur{Yang}}

\author[2]{\fnm{Riwei} \sur{Wang}}

\author*[1]{\fnm{Xuping} \sur{Zhang}}\email{xuzh@mpe.au.dk}

\affil*[1]{\orgdiv{Department of Mechanical and Production Engineering}, \state{Aarhus University}, \country{Denmark}}
\affil[2]{\orgdiv{School of Data Science and Artificial Intelligence}, \state{Wenzhou University of Technology}, \country{China}}
\abstract{Hand-eye calibration, a fundamental task in vision-based robotic systems, is commonly equipped with collaborative robots, especially for robotic applications in small and medium-sized enterprises (SMEs). Most approaches to hand-eye calibration rely on external markers or human assistance. We proposed a novel methodology that addresses the hand-eye calibration problem using the robot base as a reference, eliminating the need for external calibration objects or human intervention. Using point clouds of the robot base, a transformation matrix from the coordinate frame of the camera to the robot base is established as ``\textbf{I}=\textbf{AXB}.'' To this end, we exploit learning-based 3D detection and registration algorithms to estimate the location and orientation of the robot base. The robustness and accuracy of the method are quantified by ground-truth-based evaluation, and the accuracy result is compared with other 3D vision-based calibration methods. To assess the feasibility of our methodology, we carried out experiments utilizing a low-cost structured light scanner across varying joint configurations and groups of experiments. The proposed hand-eye calibration method achieved a translation deviation of 0.930 mm and a rotation deviation of 0.265 degrees according to the experimental results. Additionally, the 3D reconstruction experiments demonstrated a rotation error of 0.994 degrees and a position error of 1.697 mm. Moreover, our method offers the potential to be completed in 1 second, which is the fastest compared to other 3D hand-eye calibration methods. We conduct indoor 3D reconstruction and robotic grasping experiments based on our hand-eye calibration method. Related code is released at \url{github.com/leihui6/LRBO}.}

\maketitle
\section{Introduction}
Robotic solutions for production and manufacturing automation play a crucial role in SMEs. They require high flexibility, minimal space usage, and the ability to share workspace with human operators \cite{yang2023automation}. Collaborative robots (cobots) are ideal for meeting these needs due to their lightweight design, ease of programming, and flexibility. Hand-eye calibration \cite{tsai1989new, horaud1995hand, strobl2006optimal} is a fundamental and necessary task due to the essential role of cobots integrated with a vision system. It refers to estimating a relative rigid relationship between the robot end-effector (hand) and its sensor (eye), typically a camera. This relationship consists of a rotation and translation matrix, and can be used to align the hand in the robotic system with its visual perception, enabling the robot to manipulate objects in its environment accurately. Three dimension (3D) and two dimension (2D)-based vision has become increasingly prevalent in robot perception and manipulation \cite{xing2022reconstruction, li2021robot, andreff1999line, koide2019general, altan2020model, altan2020performance}, and hand-eye calibration problem has been widely studied \cite{enebuse2021comparative, jiang2022overview, enebuse2022accuracy}. In this paper, we focus on 3D-based vision in which point cloud data is only used for the hand-eye calibration problem.
\begin{figure}[htbp]
    \centering
    \subfigure[Eye-in-hand calibration]{
        \includegraphics[width=0.45\linewidth]{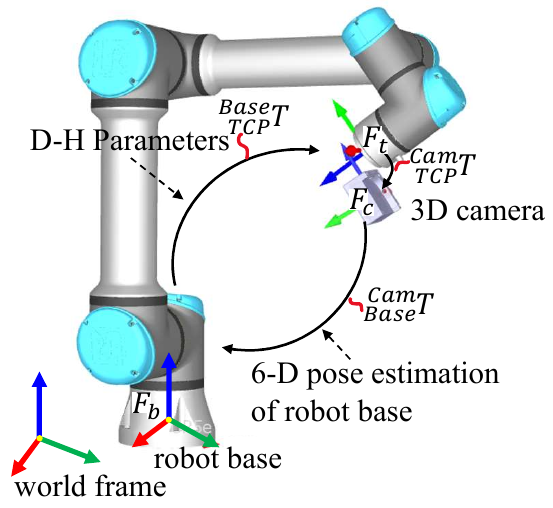}
        \label{fig.intro_eye_in_hand}
    }
    \subfigure[Eye-to-hand calibration]{
        \includegraphics[width=0.45\linewidth]{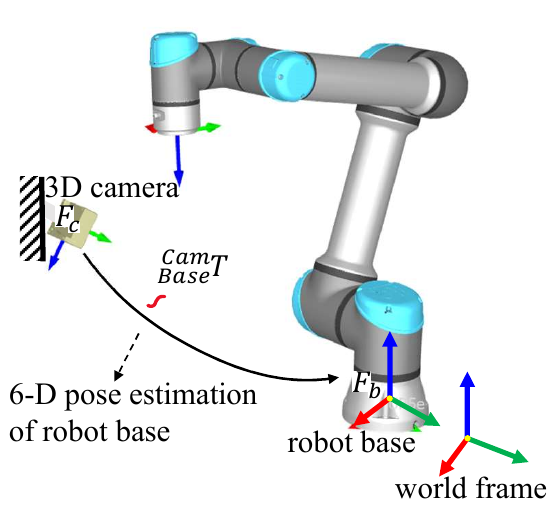}
        \label{fig.intro_eye_to_hand}
    }
    \caption{Our proposed hand-eye calibration method. We estimate the transformation matrix between the robot base and camera frame using point clouds and one robot arm movement where the robot base as an object is detected and aligned with a 3D model.}
    \label{fig.intro}
\end{figure}

In general scenarios where a collaborative robot works with a vision system, accurate calibration objects and generating multiple robot poses are always required during hand-eye calibration. For hand-eye calibration based on 2D (RGB) images, the correspondence between 3D spatial coordinates and 2D image coordinates is achieved by using some known feature points on the calibration object \cite{huang2019research}. The required calibration objects can be chessboards \cite{lee2018high} and specific labels \cite{kato1999marker, atcheson2010caltag}. More specifically, the end-effector of the robot needs to move to a series of known positions, where images and poses are collected. In the end, the collected data is used to estimate the transformation matrix. Furthermore, for approaches based on the point cloud or depth image, 3D objects are utilized, such as a 3D cube \cite{xu2017tcp} and a standard sphere \cite{huang2019research}. The existing hand-eye calibration approaches are faced with several key challenges: 1) they always require an additional calibration object; 2) they are time-consuming due to the multiple movements of the robot; 3) human support is required during calibration process. While hand-eye calibration is typically considered a one-time task, recalibrations are necessary and can be challenging for numerous robots integrated with a vision system. Moreover, the parameters of a robotic system changing over time can affect the calibrated relative position between the camera and the end-effector, such as wear and tear \cite{levine2018learning}, component replacement, environmental changes \cite{ma2018modeling}, and a specific vibration \cite{wu2021simultaneous} during movement for mobile robots. Therefore, fast and reliable hand-eye calibration/recalibration is essential for vision-guided robot systems

To overcome these drawbacks, this paper proposes a method for solving hand-eye calibration with point clouds and one robot arm movement. We use the robot base as the calibration object to eliminate the need for additional equipment, making it more efficient and out-of-the-box. Given point clouds of the robot base, the position and orientation of the robot base are estimated relative to the frame of the 3D camera. For this purpose, our method involves detecting the robot base and aligning it with a corresponding 3D model in the form of point cloud. This process can finally determine the transformation between the camera frame and the robot base, $^{Cam}_{Base}\textbf{T}$. In the case of eye-in-hand calibration, the transformation between the camera and the robot flange or more specifically, the tool center point (TCP), $^{Cam}_{TCP}\textbf{T}$, can be calculated by an inverse matrix operation, as shown in Fig. \ref{fig.intro_eye_in_hand}. In the case of eye-to-hand calibration, the transformation between the robot base and the camera frame, $^{Cam}_{Base}\textbf{T}$, can be estimated directly, as shown in Fig. \ref{fig.intro_eye_to_hand}. 

To summarize, the contributions of our work are as follows:
\begin{itemize}
    \item We present a novel, fast, and reliable hand-eye calibration approach that uses the native object, the robot base, as the calibration object and requires only one robot arm movement.
    \item We establish a learning-based 3D detection and registration framework to extract and estimate the 6D pose of the robot base. Additionally, we propose a procedure for generating low-overlap point cloud datasets in virtual environments.
    \item Comprehensive experiments are conducted to evaluate the robustness and accuracy of our proposed methods, including calibration at multiple joint configurations and 3D reconstruction experiments. Demonstrations of 3D reconstruction as a byproduct and robotic grasping tasks are presented at the end.
\end{itemize}

The remaining part of the paper is organized as follows: In Section \ref{related}, we review the related work on hand-eye calibration. The proposed method is described in detail in Section \ref{methods}. Section \ref{Implementation} presents the implementation details and evaluation for the learning-based detection and registration algorithm. In Section \ref{experiments}, we evaluate the robustness and accuracy of our method through ground-truth-based experiments. In addition, the indoor 3D reconstruction and robotic grasping experiments are conducted. Finally, Section \ref{conclusion} summarises the main conclusions and discusses future work.

\section{Related work}\label{related}
Various methods have been proposed to address the hand-eye calibration problem, ranging from traditional approaches using RGB images to more recent methods incorporating 3D data. For methods that rely on 2D images as input, the procedure typically involves capturing a series of images from different viewpoints, each containing calibration patterns. Current hand-eye calibration methods can be broadly classified into four mathematical models \cite{jiang2022overview}: ``\textbf{AX}=\textbf{XB}'' \cite{tsai1989new, shiu1987calibration}, ``\textbf{AX}=\textbf{YB}'' \cite{zhuang1994simultaneous, dornaika1998simultaneous}, ``\textbf{AXB}=\textbf{YCZ}'' \cite{wu2016simultaneous, qin2022simultaneous} and reprojection error-based calibration (REC) \cite{zhi2017simultaneous, ali2019methods, schmidt2005calibration}.

In the forms of ``\textbf{AX}=\textbf{XB}'' and ``\textbf{AX}=\textbf{YB},'' calibration objects are utilized to establish the camera-world frame. The form of ``\textbf{AXB}=\textbf{YCZ}'' is used for multiple robots, and the calibration patterns on the robot flange are required, where the checkerboard \cite{antonello2017fully}, ArUco \cite{garrido2014automatic}, ARToolKit \cite{abdullah2002camera}, and ARTag \cite{fiala2005artag} are commonly used. Previous works related to ``self-calibration'' mainly rely on REC \cite{zhi2017simultaneous, ali2019methods, schmidt2005calibration}, motion estimation \cite{lee1996self, wei1998active, heller2011structure, andreff2001robot, schmidt2005calibration}, and tool-tracking \cite{pachtrachai2016hand}. Although these methods do not rely on special calibration objects, they cannot easily be used in a non-rigid environment where the geometry of the targets is unknown \cite{pachtrachai2016hand}. Moreover, existing methods typically require a minimum of two movements to achieve calibration, and additional movements may be necessary to improve accuracy \cite{horaud1995hand, strobl2006optimal}. 

With the availability of affordable 3D cameras, an increasing number of applications are incorporating 3D sensors that can generate both depth images and point clouds. Their hand-eye calibration problem was still solved with calibration patterns. For example, a criterion sphere \cite{xu2017tcp}, 3D cube \cite{huang2019research}, the planner-like object \cite{chen2016noise} and pin-like object \cite{wagner2015self} are utilized in 3D vision-based hand-eye calibration. However, these methods require high-quality manufacturing of the calibration objects and are practically inappropriate for many application due to the complex operations for calibration.

In addition to methods that use designed calibration objects, a calibration method utilizing arbitrary objects was introduced by \cite{xing2022reconstruction} and \cite{peters2024robot}, which employs a modified iterative closest point algorithm to align the point clouds captured at different robot poses. Our proposed method is similar to theirs. However, the calibration object in our method is the robot base rather than a third-party object. In addition, instead of having many robot poses in the hand-eye calibration process, only one is needed. Specifically, we build a transformation matrix as ``\textbf{I}=\textbf{AXB},'' where $\textbf{A}$ denotes the forward kinematics whose parameters are supposed to be known, and $\textbf{B}$ is estimated by point cloud registration. $\textbf{X}$ is unknown and represents the transformation matrix between the robot flange and the camera. 

There have been some similar works aimed at estimating the pose of robotic arms. One such work was presented by \cite{lee2020camera}, where camera-to-robot pose estimation from a single RGB image is achieved. Another related approach is proposed by \cite{bohg2014robot}, which estimates the pose of a marker-less robot arm using only one frame of depth image, making one-shot hand-eye calibration possible without external calibration objects. It is worth noting that one approach, seline \cite{wong2017segicp}, estimates the transformation between the robot base and the camera frame by estimating the pose of the robot's end effector in the camera optical frame. However, these methods may not be easily adaptable to eye-in-hand calibration, as it can be difficult to capture the end-effector in the camera frame. In contrast, our proposed method can be used for eye-in-hand and eye-to-hand calibration.

\section{Problem Definition and Methodology}\label{methods}
In this section, we define the hand-eye calibration problem, including both eye-in-hand and eye-to-hand calibration, and present the classical solutions ``\textbf{AX}=\textbf{XB}'' for each. We then elaborate on our proposed hand-eye calibration based on the learning-based 3D vision.

\subsection{Hand-eye Calibration}
There are various coordinate systems in a robot system, such as the robot base, sensors and tool center point (TCP) or robot flange. In order to perform robot manipulation and build a better understanding for the surrounding environment, it is crucial to establish the rigid relationship between these coordinate systems. Hand-eye calibration is to determine the transformation matrix between the robot flange $F_t$ and the camera frame $F_c$.

To be more specific, the robot base coordinate system, denoted as $F_b$, is the reference for robot movements, as depicted in Fig. \ref{fig.intro}. The image data, such as 2D images and 3D point clouds, are collected in the camera frame $F_c$. To execute a vision-guided task, determining the transformation between $F_c$ and $F_b$ is essential. The transformation matrix from $F_t$ to $F_b$ follows the D-H model of the robot arm, while the objective of hand-eye calibration is to estimate the transformation matrix from $F_c$ to $F_t$.

In general, depending on whether the camera is mounted on a robot arm, there are two kinds of hand-eye calibration, known as eye-in-hand and eye-to-hand calibration.

\subsection{Eye-in-Hand Calibration}
Eye-in-hand calibration involves mounting cameras, either 2D or 3D, on the robot arm. The objective is to determine the rigid transformation matrix from the robot flange or TCP to the camera frame. The $_{Cam}^{TCP}\textbf{T}$ consists of a matrix of rotation ($\textbf{R}_{3\times3}$) and translation ($\textbf{t}_{3\times1}$) as
\begin{align}
        \label{eqn:tsfm}
        _{Cam}^{TCP}\textbf{T} = 
        \begin{pmatrix}
        \begin{array}{ccc}
            \textbf{R}_{3 \times 3}&\textbf{t}_{3 \times 1} \\
            \textbf{0}_{1 \times 3}&1
        \end{array}
        \end{pmatrix}
 \end{align}
To solve $_{Cam}^{TCP}\textbf{T}$, a common method ``$\textbf{AX}=\textbf{XB}$'' was presented, which use the constant transformation between the calibration object and robot base, i.e., the $_{Obj}^{Base}\textbf{T}$ is consistent. At $i^{th}$ and $i+1^{th}$ pose, calibration object in camera frame is recorded as $_{Obj}^{Cam}\textbf{T}_i$ and $_{Obj}^{Cam}\textbf{T}_{i+1}$, such that we have
\begin{equation}\label{eqn:tsfm3}
    \left\{\quad
        \begin{aligned}
        &_{TCP}^{Base}\textbf{T}_i\ _{Cam}^{TCP}\textbf{T}\ _{Obj}^{Cam}\textbf{T}_i=\ _{TCP}^{Base}\textbf{T}_{i+1}\ _{Cam}^{TCP}\textbf{T}\ _{Obj}^{Cam}\textbf{T}_{i+1}\\
        &\vdots \\
        &_{TCP}^{Base}\textbf{T}_{n-1}\ _{Cam}^{TCP}\textbf{T}\ _{Obj}^{Cam}\textbf{T}_{n-1}=\ _{TCP}^{Base}\textbf{T}_{n}\ _{Cam}^{TCP}\textbf{T}\ _{Obj}^{Cam}\textbf{T}_{n}
        \end{aligned}
    \right.
\end{equation}
where $_{Cam}^{TCP}\textbf{T}$ can be assumed as a constant value, and $n$ denoted the number of robot motion. At the $i^{th}$ and $i+1^{th}$ pose, a formula of ``\textbf{AX}=\textbf{XB}'' can be derived by
\begin{align}\label{eqn:tsfm4}
\underbrace{_{TCP}^{Base}\textbf{T}_{i+1}^{-1}{}_{TCP}^{Base}\textbf{T}_i}_{A}{}\ \underbrace{_{Cam}^{TCP}\textbf{T}}_{X}=\underbrace{_{Cam}^{TCP}\textbf{T}}_{X}\ \underbrace{_{Obj}^{Cam}\textbf{T}_{i+1}{}_{Obj}^{Cam}\textbf{T}_i^{-1}}_{B}
\end{align}
where $_{Cam}^{TCP}\textbf{T}$ can be estimated by nonlinear solver \cite{horaud1995hand}, Euclidean group \cite{park1994robot}, dual quaternions \cite{daniilidis1999hand, daniilidis1996dual}, Kronecker product \cite{andreff2001robot, wang2022optimal}, linear equations \cite{zhao2009hand} and optimized neural network \cite{hua2021hand} etc. 

However, our method simplifies the process by estimating the 6D pose of the robot base with respect to the camera frame, denoted as $_{Base}^{Cam}\textbf{T}$ in a single movement. This eliminates the need for multiple movements during the hand-eye calibration process. To be more specific, the relationship is ``\textbf{I}=\textbf{AXB}'' which can be explained as
\begin{align}\label{eqn:tsfm5}
\textbf{I}=\ _{TCP}^{Base}\textbf{T}\ _{Cam}^{TCP}\textbf{T}\ _{Base}^{Cam}\textbf{T}
\end{align}
where $_{TCP}^{Base}\textbf{T}$ can be determined through the D-H model that is supposed to be known, while the $_{Base}^{Cam}\textbf{T}$ is obtained through 6D pose estimation for the robot base in our study, as illustrated in Fig. \ref{fig.intro_eye_in_hand}. As a result, $_{Cam}^{TCP}\textbf{T}$ can be derived directly as 
\begin{align}\label{eqn:tsfm6}
_{Cam}^{TCP}\textbf{T} = _{TCP}^{Base}\textbf{T}^{-1}{}\ _{Base}^{Cam}\textbf{T}^{-1} 
\end{align}

\subsection{Eye-to-Hand Calibration}
Eye-to-hand calibration refers to a camera mounted stationary outside the robot, where the camera can have a global view of the workspace. The purpose of eye-to-hand calibration is to determine a rigid transformation matrix between the camera frame and the robot base frame ($^{Base}_{Cam}\textbf{T}$). By attaching a calibration object to the robot flange, we utilize the consistent displacement from the calibration object to the robot flange, such that we have 
\begin{equation}\label{eqn:tsfm7}
    \left\{\quad
    \begin{aligned}
    &_{Base}^{TCP}\textbf{T}_i{}\ _{Cam}^{Base}\textbf{T}\ _{Obj}^{Cam}\textbf{T}_i=\ _{Base}^{TCP}\textbf{T}_{i+1}{}\ _{Cam}^{Base}\textbf{T}\ _{Obj}^{Cam}\textbf{T}_{i+1}\\
    &\vdots\\
    &_{Base}^{TCP}\textbf{T}_{n-1}{}\ _{Cam}^{Base}\textbf{T}\ _{Obj}^{Cam}\textbf{T}_{n-1}=\ _{Base}^{TCP}\textbf{T}_{n}{}\ _{Cam}^{Base}\textbf{T}\ _{Obj}^{Cam}\textbf{T}_{n}
    \end{aligned}
    \right.
\end{equation}
where $n$ denotes the number of robot motion and $\textbf{T}_{i}$ denotes the rigid transformation corresponding to the $i^{th}$ robot motion. Similar with eye-in-hand calibration, Eq.\ref{eqn:tsfm7} also can be represented as Eq.\ref{eqn:tsfm8} at $i^{th}$ and $i+1^{th}$ pose.
\begin{align}
    \label{eqn:tsfm8}
    \underbrace{_{Base}^{TCP}\textbf{T}_{i+1}^{-1}{}\ _{Base}^{TCP}\textbf{T}_i{}}_{A}\ \underbrace{_{Cam}^{Base}\textbf{T}}_{X}&=\underbrace{_{Cam}^{Base}\textbf{T}}_{X}\ \underbrace{_{Obj}^{Cam}\textbf{T}_{i+1}{}\ _{Obj}^{Cam}\textbf{T}_i^{-1}}_{B}
\end{align}  
where $_{Base}^{TCP}\textbf{T}$ can be provided by a robotic kinematic model, and $_{Obj}^{Cam}\textbf{T}$ is obtained from the feature detection result of images. It is necessary to have at least two robot motions to uniquely solve for $\textbf{X}$. Meanwhile, a greater number of motions $n$ may be required to achieve a higher level of accuracy \cite{horaud1995hand, strobl2006optimal}. 

Our proposed also can be applied to the eye-to-hand calibration. Specifically, the rotation and translation of the robot base can be determined in the camera frame to build $_{Base}^{Cam}\textbf{\textbf{T}}$, given the captured point clouds. In addition, in practice applications, the camera is typically placed next to the robot, providing a overall view of the workspace. This allows us to determine the transformation matrix $_{Base}^{Cam}\textbf{\textbf{T}}$ from the point clouds captured in each camera shot, as illustrated in Fig. \ref{fig.intro_eye_to_hand}.

\subsection{6D-Pose Estimation of Robot Base}
The objective of eye-in-hand calibration is to estimate the transformation matrix between the frame of the camera and TCP. Our method involves estimating the 6D pose of the robot base. The goal of eye-to-hand calibration is achieved by finding the transformation matrix between the frame of the camera and the robot base. It is evident that a common step in both calibrations is the estimation of the transformation matrix between the robot base and the camera frame. This section provides a detailed explanation of the 6D pose estimation process.

\begin{figure}[htbp]
    \centering
    \includegraphics[width=1.0\linewidth]{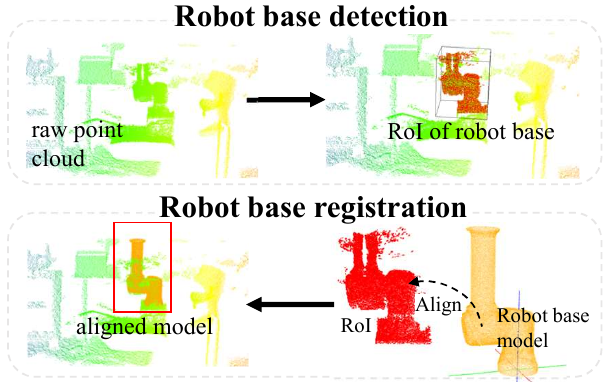}
    \caption{Our developed pipeline for the robot base 6D-Pose estimation.}
    \label{fig3.4}
\end{figure}

Our developed pipeline for robot base 6D-pose estimation is depicted in Fig. \ref{fig3.4}. The goal of robot base detection is to obtain a ROI of the robot base, which is used as a source data in the registration framework. The point cloud registration problem can be mathematically described as finding the optimal rigid transformation, $\textbf{T} = [\textbf{R}\ |\ \textbf{t}]$, that aligns a source point cloud, $P=\left\{p_i \in \mathbb{R}^{3} \mid i=1 \dots n\right\}$, with a reference point cloud, $ Q=\left\{q_j \in \mathbb{R}^{3} \right.$ $\left. \mid j=1 \dots m\right\} $, where $\textbf{R}$ is a $3\times3$ rotation matrix and $\textbf{t}$ is a $3\times1$ translation vector. The objective is to minimize the distance between corresponding points in the two point clouds as
\begin{align}
    \label{eqn:reg}
    \min_{T} \sum_{i=1}^{n} \min_{j=1}^{m} \left\|p_{i} - \textbf{R}p_{j} - \textbf{t}\right\|_2
\end{align}
where $\left\| \cdot \right\|_{2}$ is the Euclidean norm. In our study, the point cloud $P$ is the RoI of the robot base, and the point cloud $Q$ is the 3D model of the robot bast.

In summary, the pipeline consists of three steps:
\begin{enumerate}
    \item Robot Base Detection: Given point clouds, the robot base is detected with a bounding box and extracted automatically as a region of interest (RoI). 
    \item Robot Base Registration: The extracted point cloud is aligned with a reference robot base model represented as a point cloud.
    \item 6D Pose Estimation: The position and rotation matrix of the robot base are determined with respect to the camera frame.
\end{enumerate}

More specifically, since the ROI is based on the camera's frame and the reference model has its own frame, the transformation matrix between the ROI and the reference model is denoted as $^{Cam}_{Ref}\textbf{T}$, and we have
\begin{align}
    \label{eqn:cam2ref}
    ^{Cam}_{Ref}\textbf{T}\ =\ _{Ref^{\prime} }^{Cam}\textbf{T}\ _{Ref}^{Ref^{\prime}}\textbf{T}
\end{align}

where $_{Ref}^{Ref^{\prime}}\textbf{T}$ is known and consists of a series of scaling, rotation, and translation manipulations, as the raw reference model needs adjustments to align with the acquired data. $_{Ref^{\prime} }^{Cam}\textbf{T}$ is obtained from the point cloud registration framework.
A translation vector from the camera frame to the robot base frame is given by 
\begin{align}
    \label{eqn:cam2base_p}
    ^{Cam}_{Base}\textbf{t}\ =\ ^{Cam}_{Ref}\textbf{T}\cdot \textbf{P}_{0}
\end{align}
where $\textbf{P}_{0}=\{0,0,0,1\}$ is an origin point in the reference point cloud (robot base model). A vector containing the first three elements in $^{Cam}_{Base}\textbf{t}$ is considered a translational matrix. A rotation matrix of the RoI with respect to the camera frame is then obtained by 
\begin{align}
    \label{eqn:cam2base_r}
    ^{Cam}_{Base}\textbf{R}\ =\ \text{norm}\left(^{Cam}_{Ref}\textbf{R}\right) \cdot \textbf{R}_{z}\left( -\theta_{1} \right)
\end{align}
where $norm$ is the normalisation for each row in $^{Cam}_{Ref}\textbf{R}$, which is a rotation matrix of $^{Cam}_{Ref}\textbf{T}$. $\textbf{R}_{z} \left( \cdot \right)$ denotes a rotation matrix along the Z-axis, $\theta_{1}$ is the rotation angle of the first joint. Therefore, we can express the transformation matrix between the camera and robot base frame as 
\begin{align}
        \label{eqn:cam2base}
        ^{Cam}_{Base}\textbf{T} = 
        \begin{pmatrix}
        \begin{array}{ccc}
            \text{norm}\left(^{Cam}_{Ref}\textbf{R}\right) \cdot \textbf{R}_{z} \left( -\theta_{1} \right) &^{Cam}_{Ref}\textbf{T}\cdot \textbf{P}_{0} \\
            \textbf{0}_{1 \times 3}&1
        \end{array}
        \end{pmatrix}
 \end{align}


To be consistent with the 3D model, we fix the first and second joint of the robot base at 0.0 and -90.0 degrees during hand-eye calibration. Other angles are possible if they can avoid symmetry problems and capture the geometric features of the robot base. For example, 0.0 degrees for the second joint can provide alignable data. For other robots, the configuration of specific joints can be determined based on their corresponding models. In addition, relying solely on first joint data makes it difficult to identify the rotation and translation accurately, especially when dealing with cylindrical objects, where the rotation and translation matrix may be uncertain. Therefore, the use of a second joint in the robot base is essential to overcome potential symmetry-related challenges during the point cloud registration process.



A collaborative robot, UR5e is utilized in our study and its 3D model is obtained from the official Universal Robots website. It is worth noting that the coordinate system of the UR5e model should be consistent with the real UR5e robot. In other words, the origin and coordinate system of the first joint are at the bottom of the robot base. A sampling tool provided by Cloudcompare \cite{girardeau2016cloudcompare} is used to convert the 3D mesh into point cloud. 

\section{Implementation}\label{Implementation}
\begin{figure*}
    \centering
    \includegraphics[width=\linewidth,scale=1.00]{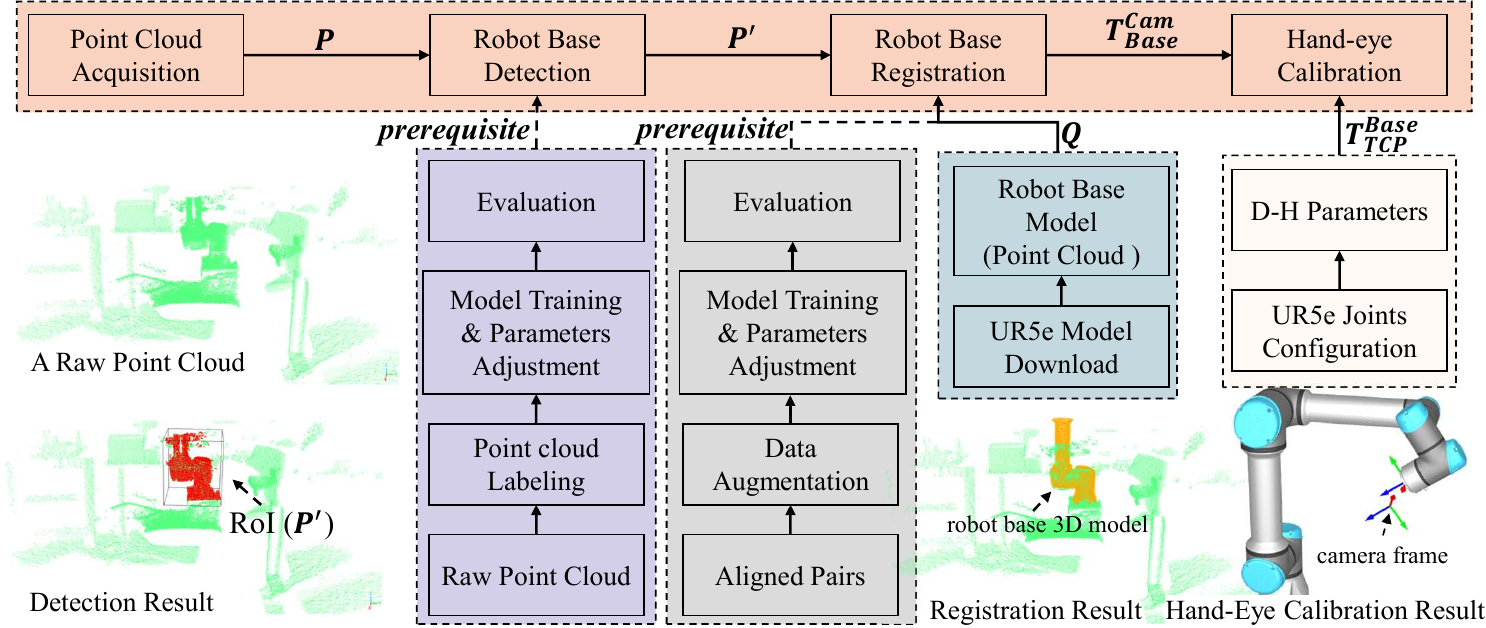}
    \caption{Overview of our proposed method. Given a point cloud ($\textbf{P}$) from a 3D camera, the robot base is detected and located as an RoI ($\textbf{P}^{'}$) via a learning-based 3D detection framework PV-RCNN++, which is trained and evaluated with our real-world dataset for the robot base task. In addition, the number of training datasets for robot base registration is increased by point cloud augmentation using a small number of aligned point clouds. We adopted a low-overlap designated learning-based framework, PREDATOR, to align the RoI with a 3D model ($\textbf{Q}$) of the robot base. The performance of the registration is also evaluated with our real data. Finally, the hand-eye calibration is solved with the registration result ($^{Cam}_{Base}\textbf{T}$) and the D-H parameter model ($^{Base}_{TCP}\textbf{T}$).}
    \label{fig:overview_workflow}
\end{figure*}
In this section, our proposed method is implemented. The developed framework is illustrated in Fig. \ref{fig:overview_workflow} where a learning-based 3D detection and registration method is introduced. 

We acquire point clouds and build our training datasets for the detection framework using a low-cost structured light camera (LIPSedge L215u), which is widely utilized in SME environments and industrial applications. As described in the manufacturer's product manual, the camera has several key features: 1) up to 0.3\% at 100cm accuracy with depth resolution 1280*800; 2) FoV 50$^\circ$ x 74$^\circ$ x 84$^\circ$ with range 0.3 to 1 m; 3) 640x480 depth output at 60fps. In addition, it produces point clouds with a minimum distance resolution of 0.9 mm at a working distance of 50 cm. The point clouds captured by our camera consists of X, Y, and Z values.

Moreover, the robot arm used in our study is the UR5e, widely employed in SME environments where it collaborates with human operators. It is designed with safety as a top priority. It has been factory calibrated, and the intrinsic parameters of the camera were determined offline. The 3D camera is mounted on a two-finger gripper attached to the robot flange.

\subsection{Robot Base Detection}
The details about the training process and the evaluation of the performance of the learning-based detection framework are presented in this section.

Robot base detection aims to identify a bounding box and extract a ROI for the robot base from the given point clouds. The reasons for using a learning-based detection framework are:
\begin{enumerate}
    \item Accuracy: It can achieve higher accuracy in identifying and localizing the robot base by leveraging large amounts of annotated data.
    \item Robustness: It is more robust to variations in the environment, such as changes in lighting, occlusions, and different robot base appearances.
    \item Adaptability: It can adapt to new types of robot bases and environments by retraining the model with new data, making them flexible for different applications.
\end{enumerate}
In our previous work \cite{zhou2022learning}, we adopted a learning-based framework PV-RCNN \cite{shi2020pv}, which is designed for accurate 3D object detection from point clouds to address the task of detecting charger stations. 
An improved and state-of-art work PV-RCNN++ \cite{shi2022pv}, which is a more practical framework for real-world applications, is exploited in our method for robot base detection. It improves the PV-RCNN \cite{shi2020pv} framework by introducing a sectorized proposal-centric keypoint sampling strategy and VectorPool aggregation module. 
The computation complexity and demanded memory resources are reduced after adjustments. The literature demonstrates that PV-RCNN++ achieved remarkable performance on large-scale open datasets such as KITTI \cite{geiger2013vision} and Waymo \cite{sun2020scalability} open dataset, which are widely used and currently the largest dataset with LiDAR point clouds for 3D object detection of autonomous driving. We adopted our raw data to the format used in the KITTI dataset.

\subsubsection{Data Preparation}
A total of 1170 point clouds of the UR5e robot base are captured and preprocessed by applying scaling and transformation operations, which enables our data trainable. Furthermore, the labels of the robot base are annotated on an online platform\footnote{https://supervise.ly/} which can also be used for other robot base labeling. We use 60\% of the total point cloud for training, while the remainder is used for evaluation. Data preparation for robot base detection is performed offline and does not take time away from hand-eye calibration in practice. 


\subsubsection{Model Training}
In order to prepare the data for training, the following parameters are set: 1) range of point cloud: [-12m, 12m], [-12m, -12m], and [-2m, 4m] along X, Y, and Z axes; 2) voxel for our point clouds: [0.5,0.5,0.15]; 3) batch size and epoch: 4 and 1000. It is tainted with 1000 epochs on a GTX3090 GPU.


\subsubsection{Evaluation}
\begin{figure}[htbp]
    \centering
    \subfigure[3D-IoU]{
        \includegraphics[width=0.45\linewidth]{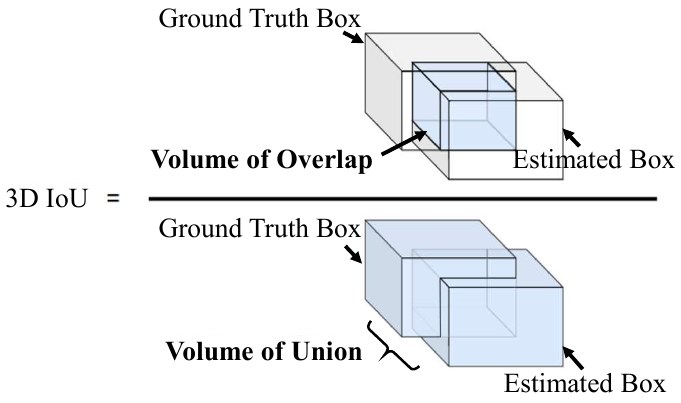}
        \label{fig:3dIoU_method}
    }
    \subfigure[An example of 3D-IoU]{
        \includegraphics[width=0.45\linewidth]{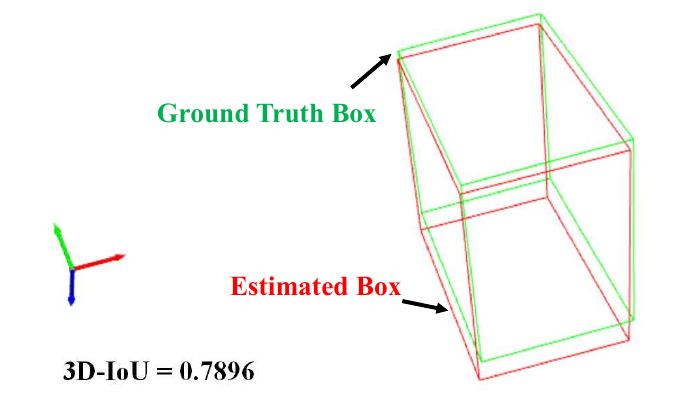}
        \label{fig:3dIoU_example}
    }
    \caption{3D-IoU and an example. An example of 3D-IoU is give on the right where the 3D-IoU is about 0.78.}
    \label{fig:3D_IoU}
\end{figure}

We evaluated our trained model using more than 400 point clouds, to test the predicted results of robot base detection in a practical scenario. Specifically, a point cloud is fed into PV-RCNN++, which outputs a predicted score and bounding box indicating the robot base. This prediction is then compared with the ground truth using 3D-IOU, a metric widely used in 3D object detection tasks \cite{shi2019pointrcnn, yang2019std}. It is determined by computing the volume of the intersection of the predicted and the ground truth bounding box, as illustrated in Fig. \ref{fig:3dIoU_method}. An example with 0.78 of 3D-IoU is given in Fig. \ref{fig:3dIoU_example}. The evaluation results are presented in Table \ref{tab:detection_evaluation} where Number indicates the size of evaluation dataset. 

\begin{figure}[htbp]
    \centering
    \subfigure[Raw point cloud]{
        \includegraphics[width=0.4\linewidth]{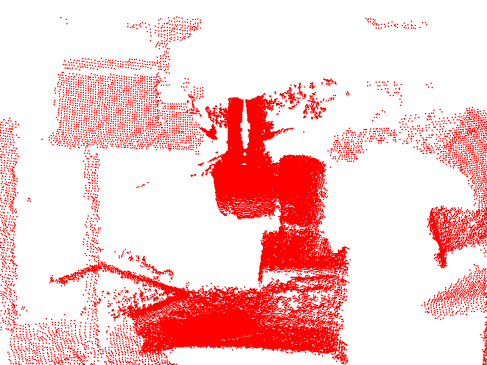}
        \label{fig5.2a}
    }
    \subfigure[Detected robot base]{
        \includegraphics[width=0.4\linewidth]{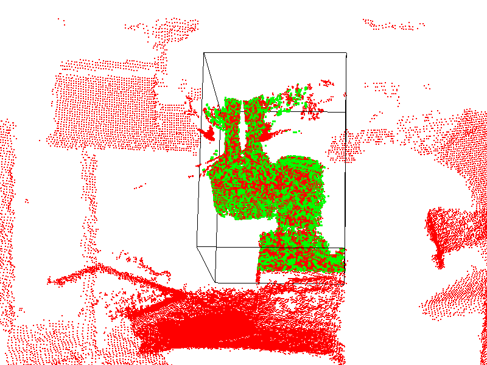}
        \label{fig5.2b}
    }
    \caption{An example of robot base detection where the robot base is shown in green.}
    \label{fig5.2}
\end{figure}

Based on the experiment results, the average 3D-IoU and score for the test dataset are found to be 0.86 and 0.96. This indicates that the prediction achieves an average overlap of 86\% with the ground truth data and a detection score of 0.96, demonstrating the stability of robot base detection. Additionally, the point cloud of the robot base is extracted as a RoI, as illustrated in Fig. \ref{fig5.2}, where the RoI is highlighted in green.

\begin{table}[]
\centering
\caption{Evaluation result of robot base detection}
\label{tab:detection_evaluation}
\begin{tabular}{@{}llll@{}}
\toprule
Item            & Standard deviation & Average & Number \\ \midrule
Predicted Score & 0.081              & 0.960   & 468    \\
3D-IoU          & 0.084              & 0.866   & 468    \\ \bottomrule
\end{tabular}
\end{table}

\subsection{Robot Base Registration}
The details on point cloud registration, including data preparation, training performance, and evaluation, are provided in this section. At the end, we present a comparison with other conventional point cloud registration methods.

Point cloud registration is challenging in our study due to the partial data of the robot base, as most registration methods are designed for nearly identical point clouds. To align the detected RoI, which is partial data, with the robot base model, we adopted a learning-based framework, PREDATOR \cite{huang2021predator}. PREDATOR is specifically designed to handle point cloud pairs with low overlap using an overlap-attention block, which uses the attention mechanism to predict overlapping regions for feature sampling by reasoning about each overlapping region. Moreover, according to the experiments reported in the literature, PREDATOR is evaluated and outperforms on 3DMatch \cite{zeng20173dmatch} and 3DLoMatch datasets, which contain scan pairs with overlaps between 10 and 30\%. In our study, the detected RoI, used as the source point cloud, represents a part of the robot base, while the reference point cloud represents the complete robot base model. 

\subsubsection{Data Preparation}
\begin{figure*}[!htp]
    \centering
    \includegraphics[width=0.9\linewidth,scale=0.8]{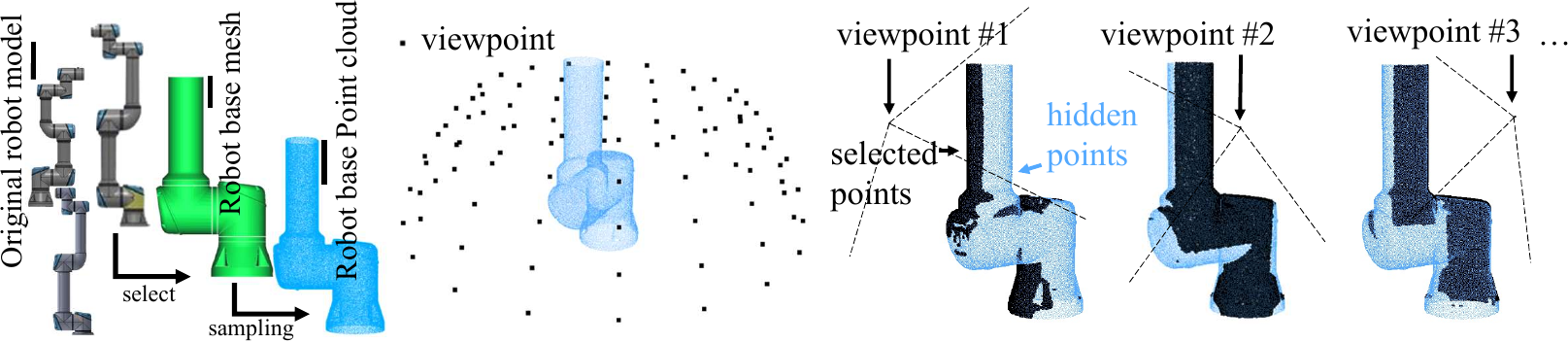}
    \caption{Robot base dataset generation. Given a point cloud of the robot base}, we simulate a number of camera positions in a virtual environment where the scanned data is estimated and selected (black). The selected data is then applied with several arbitrary transformation matrices to be training data.
    \label{fig.data_pre}
\end{figure*}

There is a lack of existing datasets for point cloud registration of the robot base, and it is challenging to ensure high quality due to the difficulty in providing accurate ground truth. To overcome these challenges, we proposed a method that uses the 3D models to generate dataset offline and accurately. Specifically, we simulated a series of viewpoints where the camera is placed and oriented towards the robot base model. The point clouds from these viewpoints are estimated and selected as source data using hidden surface estimation \cite{carpenter1984buffer}, as illustrated in Fig. \ref{fig.data_pre}. Our training and evaluation dataset includes UR3e, UR5, and UR5e, and it can also be extended to other types of robot bases given their 3D models.

\begin{figure}[htbp]
    \centering
    \includegraphics[width=\linewidth,scale=0.6]{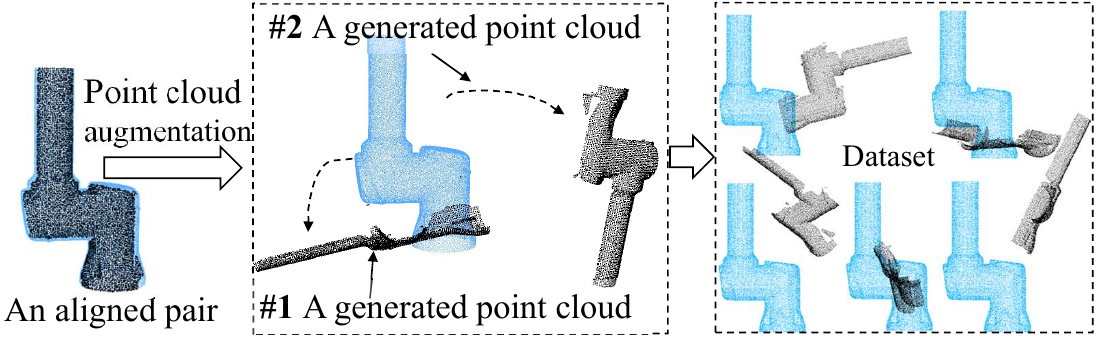}
    \caption{Point cloud augmentation in robot base registration. A point cloud on the left consists of the robot base model (black) and selected points estimated in a virtual viewpoint. After point cloud augmentation, a random transformation matrix is applied to the selected points to produce another point cloud, e.g., two point clouds in black in the middle. On the right are thousands of such point clouds in our database.}
    \label{fig:reg_dataug}
\end{figure}

Therefore, the dataset can be generated using a robot base model offline, rather than capturing data. Meanwhile, the ground truth of the transformation matrices is known and used as trainable labels. We sample 90 viewpoints within the workspace of the robot and apply arbitrary rotation and translation transformation matrices 10 times to generate a diverse and rich training dataset. The data augmentation process is shown in Fig. \ref{fig:reg_dataug}, using the UR5e as an example, where the reference point cloud is shown in blue and the generated data is shown in black.

In the dataset we built, we have 2700 pairs of aligned point clouds, including the robot base from UR3e, UR5 and UR5e. We use 60\% of the total data for training, and the remainder is used for model evaluation. 

\subsubsection{Model Training}
The PREDATOR is trained with initial learning 5e-4, momentum 0.8, and weight decay 1e-6 for 300 epochs. Additionally, the batch size is set to 1 due to memory constraints. Other parameters use default hyperparameters from the 3DMatch dataset. The best recall weight is saved and utilized for further experiments. The training process takes approximately 21 hours.


\subsubsection{Evaluation}
Given the ground truth and the estimated transformation matrix, the following metrics are used to evaluate the performance of point cloud registration for the robot base:
\begin{itemize}
    \item Root Mease Squared Error (RMSE): It \cite{li2021tutorial, huang2021comprehensive} represents the root mean square error between the estimated transformation $\textbf{T}_{est}$ and the ground truth $\textbf{T}_{gt}$. We use RMSE for rotation and translation matrix as RMSE\_T and RMSE\_R.
    
    \item Relative Rotation Error (RRE) and Relative Translation Error (RTE): RRE \cite{huang2021predator} is calculated as 
        \begin{align}
            \label{eqn:rre}
               RRE\ =\ arcos\ \left(\frac{trace\left(\textbf{R}_{est}^T \ \textbf{R}_{gt}\right)-1}{2}\right)\ 
        \end{align}
    and RTE is represented by the Euclidean distance of translation error between $\textbf{T}_{est}$ and $\textbf{T}_{gt}$.
    
    \item Rotation Error (RE): RE considers the rotation error in Euclidean norm. Given a $3\times3$ estimated rotation matrix and a ground truth matrix, we can have vector $\nu_{est}$ and $\nu_{gt}$, both of which contain $\beta$,\ $\alpha$, and $\gamma$: 
    \begin{equation}\label{eqn:RE}
        \left\{\quad
            \begin{aligned}
            &\beta=atan2\left(-R_{3,1},sqrt\left(R_{1,1}^2+R_{2,1}^2\right)\right)\\
            &\alpha=atan2\left(\frac{R_{2,1}}{cos\left(\beta\right)},\frac{R_{1,1}}{cos\left(\beta\right)}\right)\\
            &\gamma=atan2\left(\frac{R_{2,1}}{cos\left(\beta\right)},\frac{R_{3,3}}{cos\left(\beta\right)}\right)
            \end{aligned}
        \right.
    \end{equation}
    where $R_{i,j}$ indicates a element of the $i$ row and $i$ column in the rotation matrix $R$. RE is obtained as
    \begin{align}
        \label{eqn:RE2}
           RE = \| \nu_{est}-\nu_{gt} \|_{2}
    \end{align}

    \item Overlap Ratio (OR): OR considers the overlap area of the aligned point cloud and robot base model. The OR is equal to the number of matched points $N_{matched}$ divided by the total number of points of the reference point $N_{ref}$, as 
    \begin{align}
        \label{eqn:OR}
        OR\ =\ \frac{N_{matched}}{N_{ref}}
    \end{align}
    where $N_{ref}$ is a constant value. Given two point cloud $P$ and $Q$, the point in source points is considered a matched point as its distance to the nearest one is shorter than $\tau$, which is represented by $1.5 \times$ average distance in our work. The average distance is calculated as the average distance between points in the point cloud. A voxel size-based filtering algorithm is applied to ensure uniform point density for $P$ and $Q$. The ground truth OR is about 0.42 with 0.015 voxel size. 
     
\end{itemize}

The comparison experiments with traditional algorithms are conducted to determine the best algorithm based on the given metrics. These traditional algorithms include Iterative Closest Point (ICP), Go-ICP \cite{yang2015go}, FPFH-based Fast Global Registration (FGR) \cite{zhou2016fast, rusu2009fast}, 4-Points Congruent Sets Registration (4PCS) \cite{mellado2014super}, Coherent Point Drift Registration (CPD) \cite{myronenko2010point}, TEASER++ \cite{yang2020teaser}. They are widely used in the registration task due to their excellent performance in many datasets, such as synthetic range data \cite{kolluri2004spectral, curless1996volumetric}, UWA benchmark \cite{mian2006three}, 3D Match \cite{zeng20173dmatch}. 

\begin{figure}[htbp]
    \centering
    \includegraphics[width=\linewidth,scale=1.00]{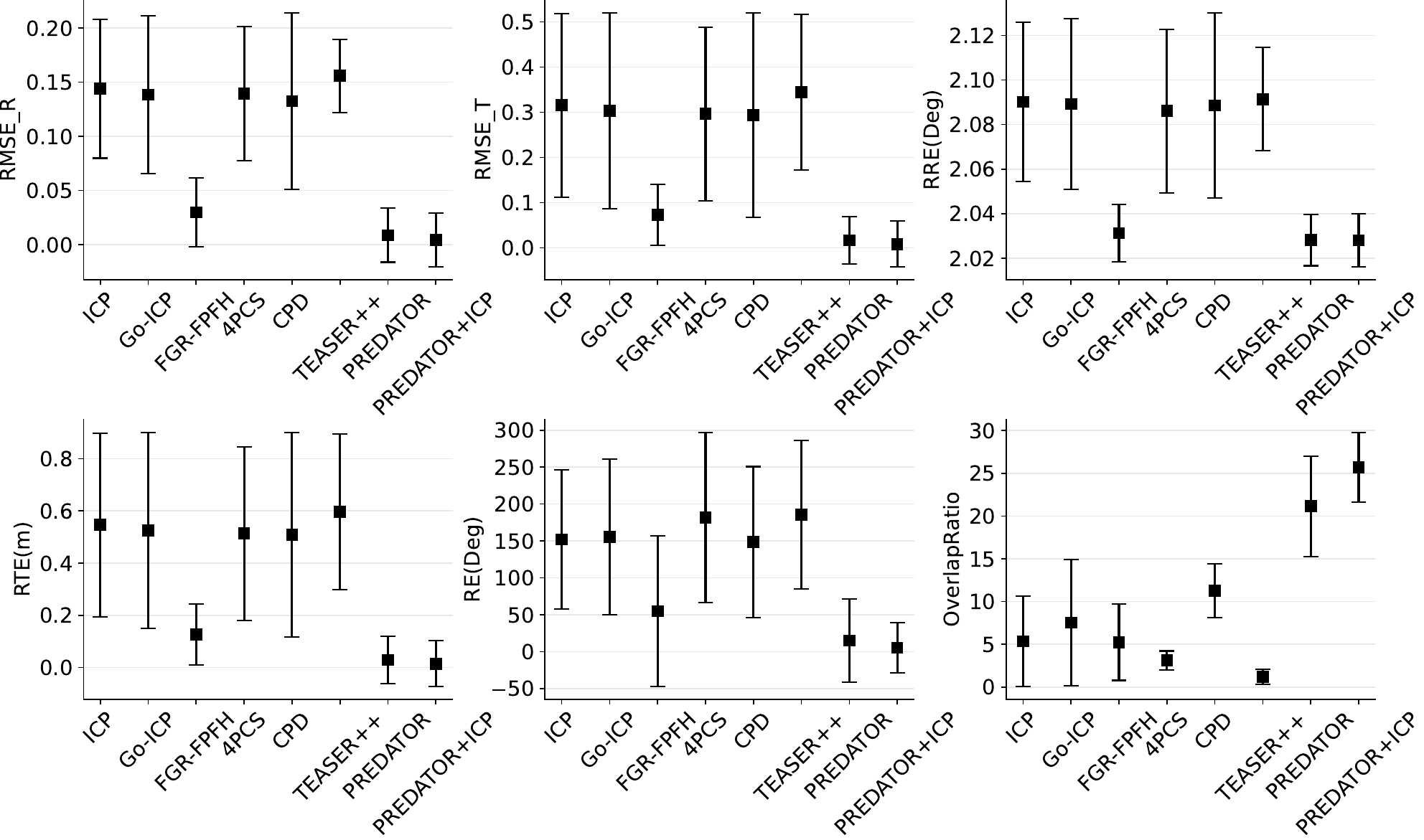}
    \caption{Compared to traditional methods. The error bars show the mean value and their lengths show the deviation. The higher the overlap ratio, the better, while the lower the other terms, the better the performance.}
    \label{fig:compwithtrad}
\end{figure}

\begin{figure*}[htbp]
    \centering
    \includegraphics[width=\linewidth,scale=1.00]{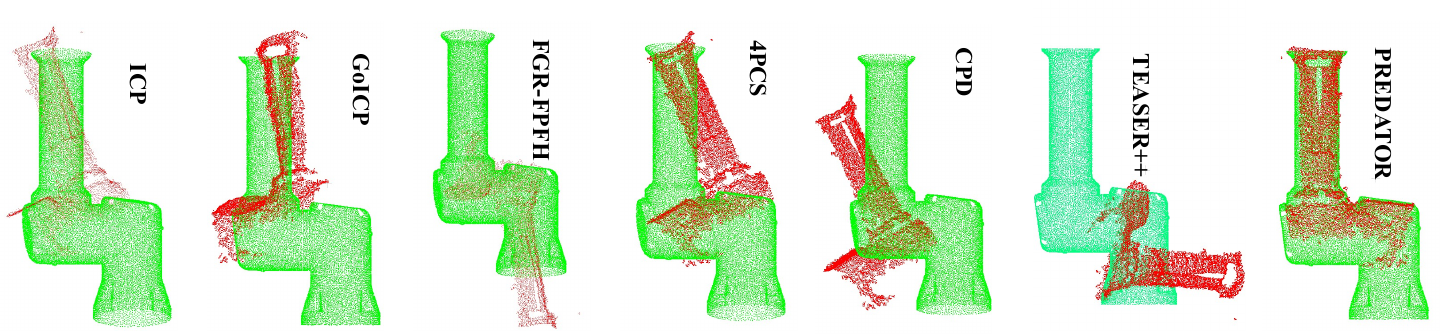}
    \caption{Examples of registration results where the scanned robot base (red) is aligned with the robot base model (green)}
    \label{fig:reg_result1}
\end{figure*}

An experiment result is shown in Fig. \ref{fig:compwithtrad}, in which the performance of different point cloud registration methods is presented. The registration result is converted into translation and rotation along the X, Y, and Z axes. Error bars represent the standard deviation and mean value. According to the results, PREDATOR exhibits a minor deviation with the largest overlap, indicating higher reliability. Moreover, the experiments confirm that PREDATOR integrated with a ICP algorithm can refine the registration result in terms of RMSE, RMSE, RTE, RE and OR. An example of the robot base registration result provided by these methods is shown in Fig. \ref{fig:reg_result1}.

In addition, the experimental results are summarised in Table \ref{tab:reg}, where the means of each metric are presented, with the best value highlighted in bold. SPS denotes the size of the source points, while TPS denotes the size of the target or reference points. Moreover, the runtime of the algorithms is measured on an Intel i7-10510U platform, while TEASER++ and PREDATOR are run on Ubuntu 18.04 with an Intel i9-10850K and an NVIDIA RTX3090. The ICP refinement integrated into PREDATOR benefits from faster computing resources. To accelerate the CPD registration algorithm, a voxel-based filter is applied to reduce the number of points. 

The experiments demonstrate that PREDATOR achieved impressive results on our real-world dataset, outperforming conventional approaches. For example, the registration transformation provided by PREDATOR has an average offset of 0.004 degrees, 8mm, 2.028 degree, 3mm, 5.283 degree, and 25.695\% from the ground truth in terms of RMSE\_R, RMSE\_T, RRE, RTE, RE, and OR. This indicates that PREDATOR yields the smallest offset from the ground truth transformation matrix and the largest overlap ratio with the reference data. Moreover, the results show that the combination of the PREDATOR and ICP algorithms yields a more accurate transformation, achieved in approximately 0.5 seconds. In the following study, we utilize PREDATOR with ICP to accomplish the robot base registration task.

\begin{table*}[htbp]
\centering
\caption{The performance comparison of non-learning methods and PREDATOR. $\downarrow$: low is better, $\uparrow$: high is better.}
\label{tab:reg}
\scalebox{0.85}{
\begin{tabular}{@{}llllllllll@{}}
\toprule
Method       & $\downarrow$ RMSE[R] & $\downarrow$ RMSE[T] & $\downarrow$ RRE(Deg) & $\downarrow$ RTE(m) & $\downarrow$ RE(Deg) & $\uparrow$ OR(\%) & SPS  & TPS & $\downarrow$ T(s)  \\ \midrule
ICP          & 0.144          & 0.315          & 2.09           & 0.109          & 152.058        & 5.356           & 3364 & 12940 & \textbf{0.137} \\
GOICP        & 0.138          & 0.303          & 2.089          & 0.105          & 155.189        & 7.543           & 3364 & 12940 & 3.077 \\
FGR-FPFH     & 0.03           & 0.073          & 2.031          & 0.025          & 54.777         & 5.234           & 3364 & 12940 & 1.401 \\
4PCS         & 0.14           & 0.296          & 2.086          & 0.103          & 181.776        & 3.105           & 3364 & 12940 & 1.092 \\
CPD          & 0.133          & 0.294          & 2.089          & 0.102          & 148.577        & 11.261          & 630  & 1694  & 6.569 \\
TEASER++     & 0.156          & 0.344          & 2.091          & 0.119          & 185.603        & 1.215           & 3359 & 12939 & 0.749 \\
PREDATOR     & 0.009          & 0.017          & \textbf{2.028} & 0.006          & 15.058         & 21.145          & 3358 & 12838 & 0.505 \\
PREDATOR+ICP & \textbf{0.004} & \textbf{0.008} & \textbf{2.028} & \textbf{0.003} & \textbf{5.283} & \textbf{25.695} & 3358 & 12857 & 0.514 \\ \bottomrule
\end{tabular}
}
\end{table*}

\subsubsection{Discussion}

According to the study \cite{pomerleau2015review, li2021tutorial, huang2021comprehensive} on point cloud registration, traditional methods cannot achieve the same level of performance as PREDATOR for the following reasons:
\begin{enumerate}
    \item Partial Data: Aligning the model robot base with only a partial point cloud can lead to local optima. This may affect the CPD registration results as it assumes local isotropy and does not consider structural data information. 
    \item Free Features: The scanned point cloud, a partial point cloud of the robot base, is missing more representative features, which results in a lack of point correspondence obtained from the FPFH feature.
    \item Poor Quality Data: The captured data is low resolution or noise due to imaging error of 3D camera. It is challenging for 4PCS, which is sensitive to the number of sample points used and its sampling method. 
    \item Arbitrary Position and Orientation: Large difference in position and orientation exists in our testing data. It probably impacts the FGR, ICP, and Go-ICP, which are sensitive to initial alignment.
\end{enumerate}

According to the analysis of the registration performance results, the learning-based approach achieves better alignment than non-learning algorithms for partial data registration. Partial data often lacks geometric shapes and predefined statistical features, which are crucial for most non-learning methods. In addition, the learning-based approach is generalized, robust to noise, and trainable, providing better generalization capabilities.

\subsection{Hand-eye Calibration}
In the previous sections, two key modules are implemented and evaluated: 1) a 3D detection framework, PV-RCNN++, detects and provides a high-confidence ROI; 2) a learning-based method, PREDATOR, which aligns the RoI with a robot base model, provides reliable alignment and is refined with an ICP algorithm.

\begin{figure}[htbp]
    \centering
    \includegraphics[width=0.9\linewidth]{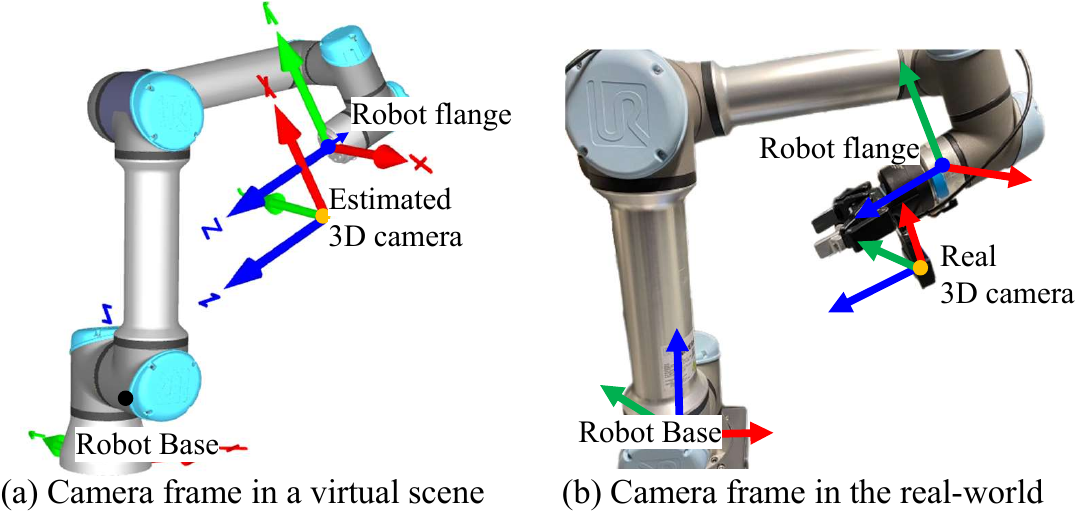}
    \caption{A real case of eye-in-hand calibration: (a) Camera frame estimated by our method, shown in a virtual scene. (b) Real camera frame in the real world.}
    \label{fig:handeyeresult}
\end{figure}

Given the transformation matrix between the camera and the robot base frame, the hand-eye calibration problem is solved by Eq. \ref{eqn:tsfm6}. A calibrated case is shown in Fig. \ref{fig:handeyeresult}, where each frame is indicated by labels such as ``Robot Base'' and ``3D Camera''. The results show that the position and orientation of the 3D camera is consistent with the real case. 

\section{Evaluation and Discussion}\label{experiments}
This section aims to evaluate the reliability and precision of our proposed hand-eye calibration method. It should be noted that it is challenging to assess the absolute precision of hand-eye calibration, as the displacement between the camera and the end-effector is difficult to measure with equipment. \cite{enebuse2021comparative}. In addition, applying experimental hand-eye calibration results, which may contain errors or large deviations, to a real robot and interacting with the real environment in practical tasks can potentially cause damage. Therefore, 3D reconstruction is relied \cite{jiang2022overview, li2021robot} upon to verify relative accuracy and evaluate the performance of our proposed method.

\subsection{Robustness Evaluation}
In this section, we evaluate the robustness of our method by performing hand-eye calibration with various suitable joint configurations, rather than restricting it to specific configurations of the robot arm. The eye-in-hand calibration is considered a case study due to its higher complexity compared to eye-to-hand calibration. In eye-to-hand calibration, the calibration results are directly provided by the 6D pose estimation of the robot base, while eye-in-hand calibration involves the use of a DH kinematics model.

\begin{figure}[htbp]
    \centering
    \includegraphics[width=0.8\linewidth]{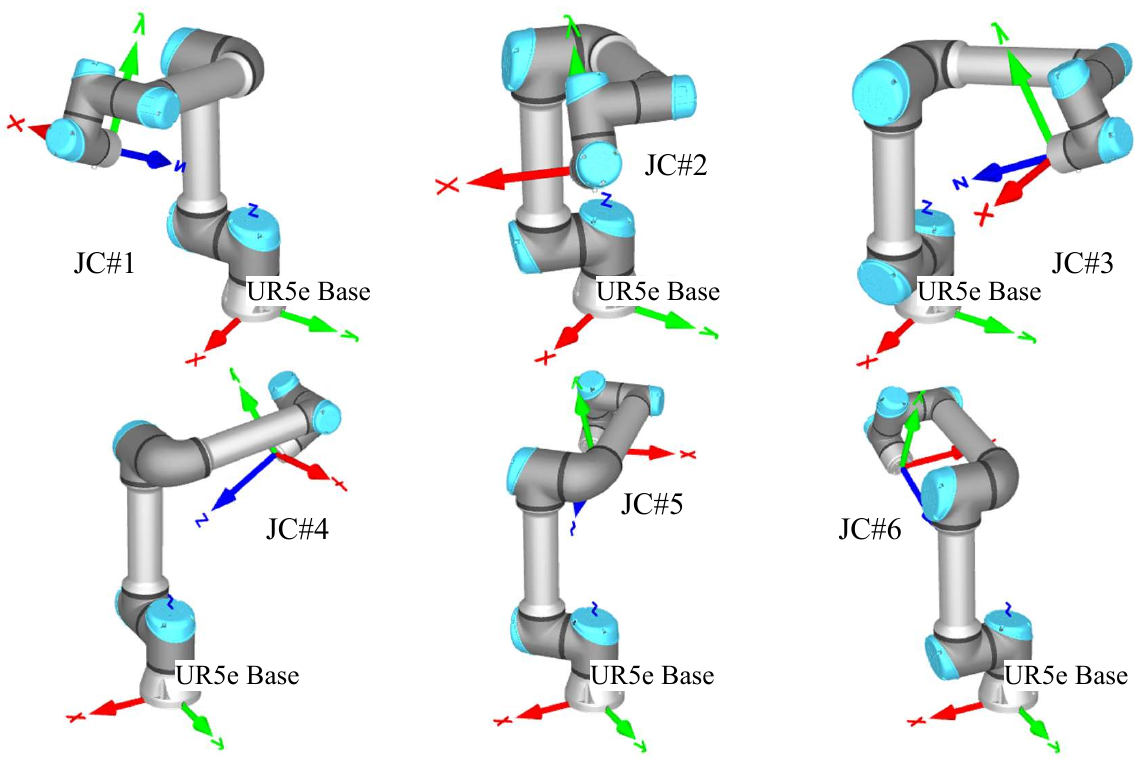}
    \caption{Six joint configurations for hand-eye calibration in which a UR5e is used as an example.}
    \label{fig:six_joints}
\end{figure}

\begin{figure}[htbp]
    \centering
    \includegraphics[width=0.9\linewidth]{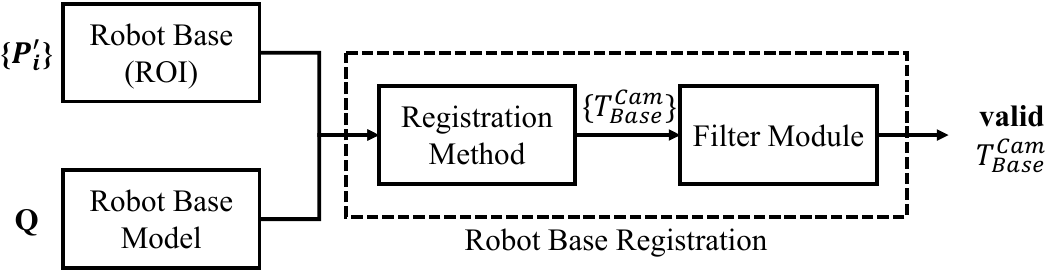}
    \caption{A filter module in the robot base registration where the transformation matrices having large overlap ratio are collected.}
    \label{fig:filter_module}
\end{figure}

To maintain consistency with the 3D model of the robot base, we set the robot's second joint to a -90.0 degree angle. This angle aligns with the 3D model, where the second joint is also at -90.0 degrees. We acquire 25 point clouds for each joint configuration in about 0.5 seconds, and subsequently apply a filtering module to derive an averaged calibration result. These results are then compared against calculations performed at six different joint configurations, as illustrated in Fig. \ref{fig:six_joints}. The filtering process is illustrated in Fig. \ref{fig:filter_module}.


These joint configurations were chosen based on two key criteria that we also recommend for optimizing other robotic systems: 1) ensuring that the robot base is within the camera's recommended working distance, and 2) including data from the first and second joints of the robot base in the scans to avoid symmetry issues during registration.

\begin{figure}[htbp]
    \centering
    \subfigure[Hand-eye calibration result calculated under six joint configurations using UR5e. The final transformation matrix from the robot flange to the robot base is represented by the translation and rotation along the X, Y and Z axes. Ideally, each value should be the same under six joint configurations, but may be different for three groups.]{
        \includegraphics[width=0.9\linewidth]{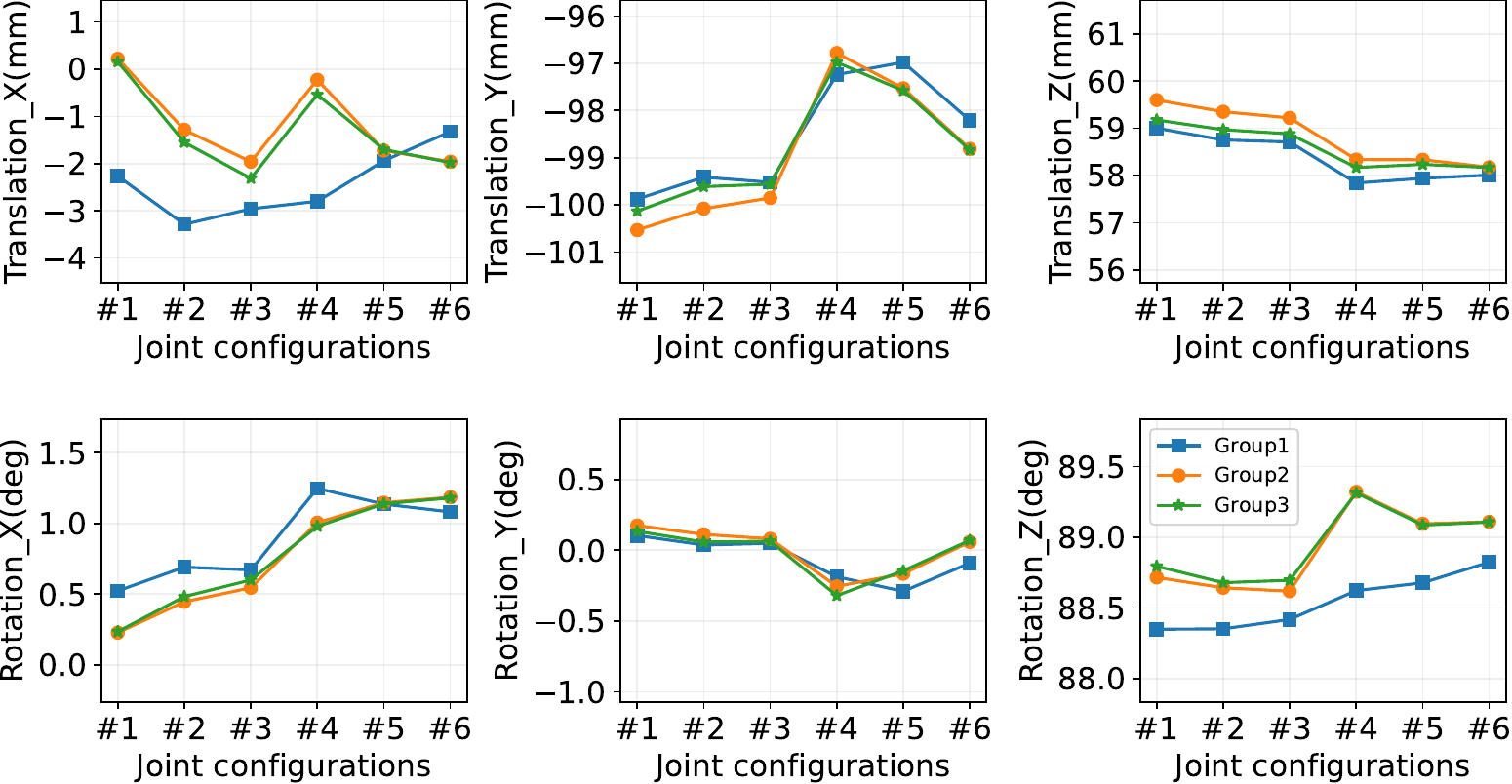}
        \label{fig:six_joints_result_1}
    }
    \subfigure[Mean value and standard deviation of the calibration results.]{
        \includegraphics[width=0.9\linewidth]{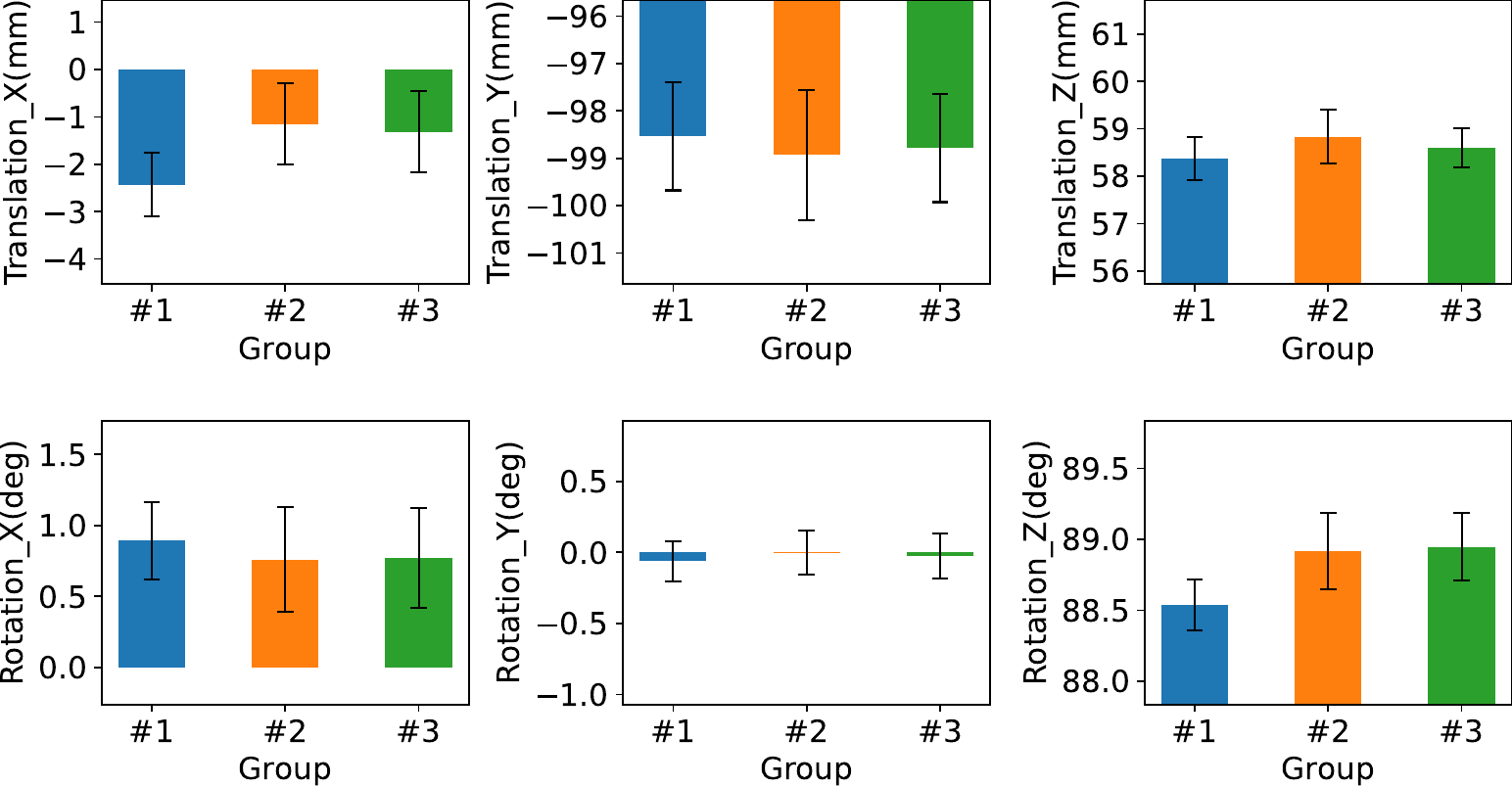}
        \label{fig:six_joints_result_2}
    }
    \caption{Testing error for three groups of experiments. Calibration results including translation and rotation along X, Y and Z axes are estimated under six joint configurations for three groups.}
    \label{fig:handeyeresult_data}
\end{figure}

We conduct hand-eye calibration experiments at six different joint configurations three times, named group1, group2, and group3. Each time involves camera reinstallation to create slight differences. The experimental results are presented in Fig. \ref{fig:handeyeresult_data} in which the results are represented as Euler angles and translations along the X, Y, and Z axes. Ideally, within a single group, the results should be identical at different joint configurations since the camera is fixed to the robot arm. However, for the three groups, slight offsets in position and orientation may exist due to camera re-mounting.

According to the results shown in Fig. \ref{fig:six_joints_result_1}, the maximum RV of translation along the X, Y and Z axes are 2.471 mm, 3.751 mm and 1.423 mm, which occur in group 1 and 2. The maximum RV of rotation along the X, Y and Z axes are 0.957 degrees, 0.454 degrees and 0.702 degrees, occurring in groups 1 and 3. The average RV is 2.548 mm for translation and 0.704 degrees for rotation based on the three groups. This indicates that the calibration results estimated at six joint configurations are very close. Furthermore, according to the results in Fig. \ref{fig:six_joints_result_2}, the maximum SD of translation of the X, Y and Z axes occurs in groups 3, 2 and 2. The maximum SD of rotation of the X, Y and Z axes occurs in groups 2, 3 and 2. The average deviation is 0.930 mm for translation and 0.265 degrees for rotation. 

Therefore, considering camera imaging and the point cloud registration error, the estimated calibration result is deemed acceptable, demonstrating the robustness of our proposed method for different joint configurations. Additionally, while a single shot can yield a calibration result, conducting multiple acquisitions is beneficial for obtaining a robust and accurate calibration, given the fast acquisition process.

\subsection{Accuracy Evaluation}
In this section, we evaluate the accuracy of the hand-eye calibration using 3D reconstructions of a plane and hemisphere. To minimize the impact of color\cite{nahler2020quantitative, sezer2021detection}, the objects used in our study were printed in white.



To obtain an accurate 3D reconstruction of the results, two types of experiments are conducted:
\begin{enumerate}
    \item Static test: It primarily estimates camera imaging errors. Specifically, we scan the object on a table without any transformations to assess fluctuations in camera imaging on the object's surfaces, including both the plane and hemisphere.
    \item Dynamic test: It primarily estimates our calibration error. Specifically, we convert the captured points to the robot base frame using our calibration result. Additionally, the transformed point clouds are refined through registration, providing the ground truth for the 3D reconstruction called Dynamic Test with ICP.
\end{enumerate}
In the 3D reconstruction experiments, points from the plane and hemisphere are manually extracted. The data acquisition distance is within the recommended range of 50 cm to 60 cm. The surfaces of the plane and hemisphere are estimated using the least squares method.

\subsubsection{Plane Reconstruction}
\begin{figure}[htbp]
    \centering
    \subfigure[The plane reconstruction setup where a cuboid object is placed on the table. The point cloud obtained at current pose if shown on the right.]{
    \includegraphics[width=0.9\linewidth]{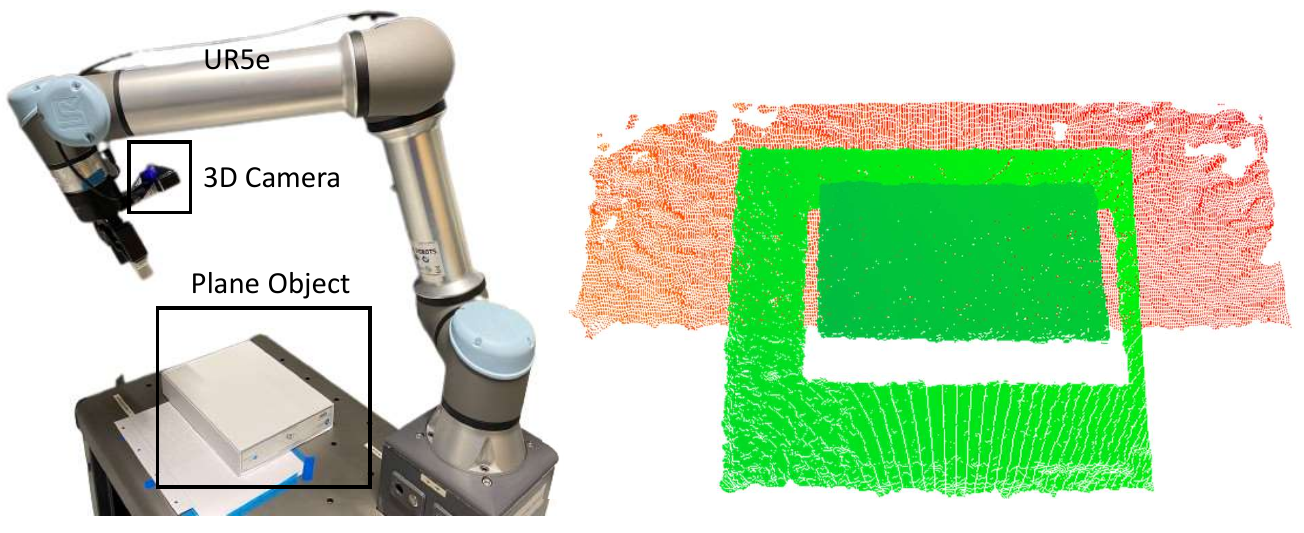}
    \label{fig:boxrecon_setup}
    }
    \subfigure[The result of plane reconstruction. The uncalibrated point clouds are on the left, the calibrated point clouds are shown on the right.]{
    \includegraphics[width=0.9\linewidth]{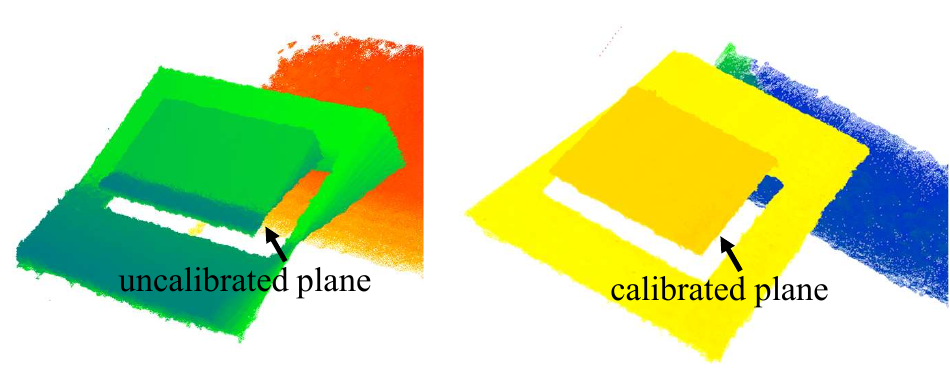}
    \label{fig:plane_recon}
    }
    \caption{Plane reconstruction}
\end{figure}

First, in static test, a cuboid is placed on the table to be scanned, with a total of 40 point clouds collected. To determine the imaging system deviation, the plane of a cuboid is scanned and the angle between the normal vectors and the RMS is estimated. The experimental setup is illustrated in Fig. \ref{fig:boxrecon_setup}, where a point cloud is shown as an example on the right.

Second, in dynamic test, we acquired 25 point clouds from various viewpoints by rotating the 3D camera around the cuboid at a constant working distance. Since these point clouds are captured from different poses, they could not be fused into a complete scene, as shown in Fig. \ref{fig:plane_recon}. In contrast, the calibrated plane point clouds form a complete and easily distinguishable scene.

The angle between normal vectors and RMS are used as metrics in the plane reconstruction. RMS is defined as
\begin{align}
    \label{eqn:RMS}
       RMS=\frac{1}{n}\sum_{i}^{n}{\parallel p_i-p_j \parallel ^2}, s.t. \parallel p_i-p_j \parallel ^2 < \tau
\end{align}
where $\tau$ is given as $1.5 \times$ average distance of the acquired point cloud. The angle is calculated by the normal vector of the top surface and the Z-axis. Ideally, both the RMS and angle are zero. 

The result of static and dynamic test are shown in Table \ref{tab:boxrecon} which includes the range of values (RV), standard deviation (SD), median and mean value. The static test results show that the camera has random noise on the surface of the object, and the average angle and RMS are 0.794 degrees and 1.006 mm. It can be considered as the imaging error during hand-eye calibration. The dynamic test results are presented along with a comparison to the static test. Moreover, an ICP registration algorithm is applied to the point clouds transformed to the robot base frame. It establishes a ground truth for the 3D reconstruction in the dynamic test, namely Dynamic test with ICP.

According to the results, the plane points show an offset of 0.994 degrees and a difference of 0.177 mm in terms of angle and RMS due to imaging errors, indicating the rotation error of our calibration result. With an offset of 0.960 mm observed in the dynamic test with ICP, the estimated rotation error is considered reliable.

\begin{table*}[]
\centering
\caption{Comparison of static and Dynamic tests of the plane reconstruction}
\label{tab:boxrecon}
\scalebox{0.9}{
\begin{tabular}{lllllll} 
\toprule
\multirow{2}{*}{} & \multicolumn{2}{l}{Static test} & \multicolumn{2}{l}{\begin{tabular}[c]{@{}l@{}}Dynamic test/\\ Dynamic test with ICP\end{tabular}} & \multicolumn{2}{l}{ $\lvert$ Offset $\rvert$ }  \\ 
                   & Angle (deg) & RMS (mm)          & Angle (deg) & RMS (mm)           & Angle (deg) & RMS (mm)          \\ \cmidrule(l){2-7} 
Range of values    & 0.731       & 1.190             & 0.599 / 0.833       & 1.097 / 0.772             & 0.132 / 0.102            & 0.093 / 0.418             \\
Standard deviation & 0.223       & 0.237             & 0.144 / 0.210       & 0.305 / 0.261             & 0.079 / 0.013            & 0.068 / 0.024             \\
Median             & 0.859       & 0.970             & 1.786 / 1.753       & 1.126 / 1.459             & 0.927 / 0.894            & 0.156 / 0.489             \\
Mean               & 0.794       & 1.006             & 1.788 / 1.754       & 1.182 / 1.505             & \textbf{0.994 / 0.960}   & 0.177 / 0.499             \\
\bottomrule
\end{tabular}}
\end{table*}

\subsubsection{Hemisphere Reconstruction}
\begin{figure}[htbp]
    \centering
    \subfigure[The setup of hemisphere reconstruction where a hemisphere object is placed on the table. The point cloud captured at current pose of robot arm is shown on the right.]{
    \includegraphics[width=0.9\linewidth]{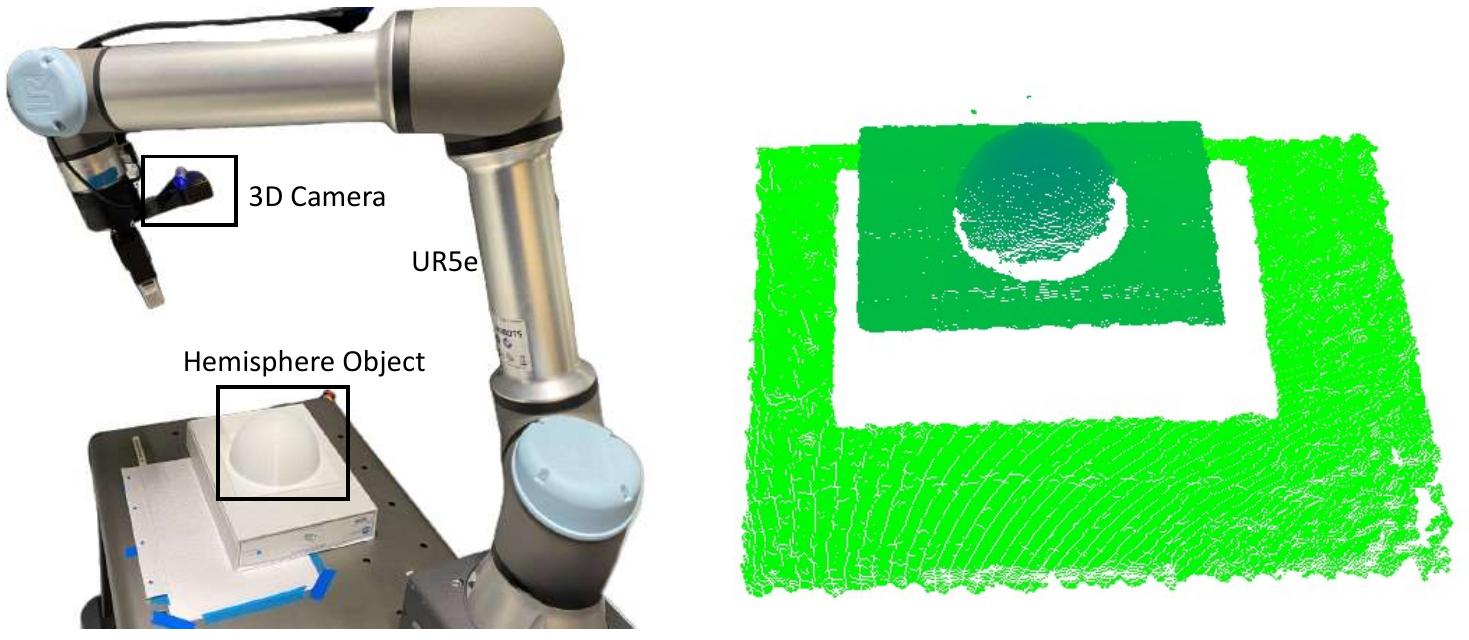}
    \label{fig:hemi_recon_setup}
    }
    \subfigure[The result of hemisphere reconstruction. The uncalibrated point clouds are on the left, while the calibrated point clouds are shown on the right.]{
    \includegraphics[width=0.9\linewidth]{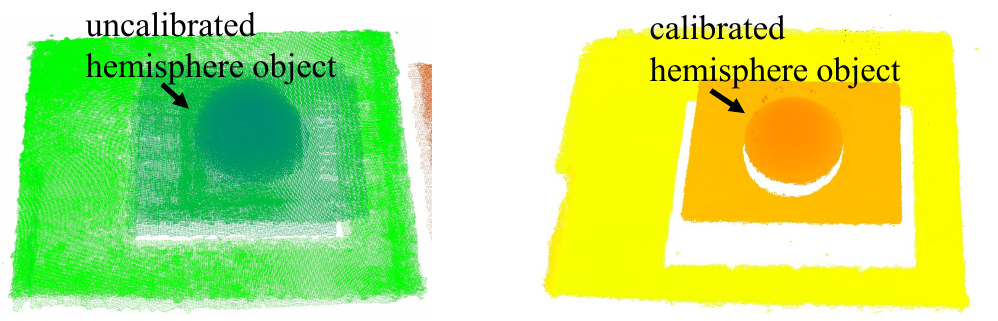}
    \label{fig:hemisphere_recon}
    }
    \caption{Hemisphere reconstruction}
\end{figure}

First, in the static test, a hemisphere placed on the table is scanned by a 3D camera without moving, collecting a total of 40 point clouds, as shown in Fig. \ref{fig:hemi_recon_setup}. 

Second, in dynamic test, a total of 40 point clouds are captured from various perspectives as the 3D camera rotates around the center of the hemisphere, with a constant working distance. The calibrated point clouds are then converted into the base frame to reconstruct the complete hemisphere object, as shown in Fig. \ref{fig:hemisphere_recon}. Considering the hemisphere shape of the object, metrics such as distance of center points (DC), estimated radius, and RMS are used to evaluate the reconstruction results. Ideally, the DC and RMS should be zero, and the designed radius for the hemisphere is 75 mm.

The static and dynamic test results are presented in Table \ref{tab:hemisphererecon}, where includes the estimated mean center point along the X, Y, and Z axes. These values are expected to be identical in the static test as the hemisphere object remains stationary. In addition, the RV, SD, median and mean are also provided to show the distribution of the experimental results.

Given the results, the average radius, DC and RMS are estimated to be 72.707 mm, 0.095 mm, and 0.694 mm according to the static test, reflecting the average imaging quality on the hemisphere surface. In addition, the conclusions drawn from the 3D hemisphere reconstruction experiment are as follows:
\begin{enumerate}
    \item The calibration result exhibits an offset of 1.046 mm, 0.754 mm and 0.517 mm in DC, radius, and RMS as compared to the results obtained from static test. This offset falls within the acceptable range, considering the 3 mm error from the imaging system.
    \item The offsets of 1.159mm, 0.697mm, and 1.025mm along the X, Y, and Z axes indicate a position error of 1.697 mm, attributed to errors from the imaging system.
    \item The offsets are 0.129 mm, 0.816 mm, and 1.146 mm along the X, Y, and Z axes, given the offset results obtained from the dynamic test with ICP and the static test. These values indicate a position error of 1.412 mm, establishing the ground truth for the 3D reconstitution.
\end{enumerate} 
Therefore, the position error of 1.697 mm is reliable based on the established ground truth of the 3D reconstruction, which is very close to the result of the dynamic test. 

In summary, based on our proposed hand-eye calibration, the 3D reconstruction experiments demonstrate an accuracy of approximately 0.994 degrees for rotation and 1.697 mm for position offset. In practical use, this offset indicates the difference between the camera frame calculated by our method and the actual camera frame position.

\begin{table*}
\centering
\caption{Comparison of static and Dynamic tests in the hemisphere reconstruction (mm)}
\label{tab:hemisphererecon}
\scalebox{0.9}{
\begin{tabular}{llllll}
\toprule
{}                     & Item                     & RV & SD & Median   & Mean     \\ \cmidrule(l){2-6}  
\multirow{6}{*}{Static test}  & Center point(x)          & 0.125       & 0.032              &   97.436   & 97.450  \\
                              & Center point(y)          & 0.087       & 0.020              & -464.726 & -464.732\\
                              & Center point(z)          & 0.272       & 0.063              &  -15.342  & -15.344 \\
                              & Radius                   & 0.158       & 0.034              &   72.709   & 72.707  \\
                              & RMS                      & 0.556       & 0.139              &    0.733    & 0.694   \\
                              & DC                       & 0.207       & 0.050              &    0.083    & 0.095   \\ \cmidrule(l){2-6} 
\multirow{6}{*}{\begin{tabular}[c]{@{}l@{}}Dynamic test/\\ Dynamic test with ICP\end{tabular}} 
                              & Center point(x)          & 2.320 / 4.185   & 0.599 / 1.088      &   96.089  / 97.150	   & 96.291  / 97.321   \\
                              & Center point(y)          & 1.516 / 2.430   & 0.434 / 0.775      & -464.095  / -464.062  & -464.035 / -463.915 \\
                              & Center point(z)          & 2.274 / 2.530   & 0.503 / 0.669      &  -14.335  / -14.261	  & -14.319  / -14.199  \\
                              & Radius                   & 1.176 / 1.066   & 0.267 / 0.281      &   71.940  / 71.970	   & 71.953  / 71.933    \\
                              & RMS                      & 0.482 / 0.489   & 0.129 / 0.144      &    1.230  / 1.304	   & 1.210   / 1.346    \\
                              & DC                       & 1.965 / 2.218   & 0.482 / 0.765      &    1.156  / 2.111	   & 1.140   / 1.959     \\ \cmidrule(l){2-6} 
\multirow{6}{*}{$\lvert$ Offset $\rvert$}       
                              & Center point(x)          & 2.196 / 4.060          & 0.567 / 1.056            & 1.348 / 0.287   & \textbf{1.159} / \textbf{0.129}   \\
                              & Center point(y)          & 1.429 / 2.344          & 0.415 / 0.756            & 0.631 / 0.664   & \textbf{0.697} / \textbf{0.816}   \\
                              & Center point(z)          & 2.002 / 2.258          & 0.440 / 0.606            & 1.008 / 1.082   & \textbf{1.025} / \textbf{1.146}   \\
                              & Radius                   & 1.018 / 0.908          & 0.233 / 0.248            & 0.769 / 0.739   & 0.754        / 0.774   \\
                              & RMS                      & 0.074 / 0.068          & 0.010 / 0.005            & 0.497 / 0.571   & 0.517        / 0.652   \\
                              & DC                       & 1.758 / 2.011          & 0.432 / 0.716            & 1.073 / 2.027   & 1.046        / 1.865   \\ \bottomrule
\end{tabular}}
\end{table*}


\subsection{Comparison and Discussion} 
In this section, we compare the position and rotation errors as well as the runtime with other conventional hand-eye calibration methods based on 3D vision. A discussion about our proposed method is provided at the end.

These methods utilize point clouds and have demonstrated explicit and excellent hand-eye calibration results. For example, Shiyu \cite{xing2022reconstruction} introduced a hand-eye calibration method for 3D-sensors that utilizes arbitrary objects. It achieves the simultaneous registration of multiview point clouds under the robot base frame through iterative refinement of the hand-eye relation. The performance of eye-in-hand calibration was evaluated by the 3D reconsecration for a known object with CAD model. Zhe \cite{wang2020efficient} proposed an eye-in-hand system, which employs a line structured light vision sensor and low-cost planar checkerboard. A pose measurement experiment was conducted to measure the quality of the calibration transformation matrix. Lei \cite{zhang2023novel} proposed a new method that uses a single sphere as a calibrator, solving the hand-eye calibration problem with high precision and noise resistance, significantly reducing fitting errors. The repeated position accuracy of the system after calibration was assessed using 3D reconstruction error and distance measurement error.

Peter \cite{peters2024robot} introduced a new approach for robot calibration using only a 3D depth sensor, treating it as an offline SLAM problem to estimate the robot's kinematic model parameters autonomously. It is compared to that of a dedicated external tracking system, showcasing comparable results at a fraction of the cost. Mingyang \cite{li2021robot} proposed a 3D reconstruction-based robot line laser hand-eye calibration method to reduce errors in point cloud reconstruction. Its evaluation was done by assessing the RMSE of the reconstructed point cloud with the proposed method. Murali \cite{murali2021situ} presented a novel method for extracting the translation component of the calibration matrix in a target-agnostic manner. The accuracy of the proposed method was evaluated by comparing the extracted translation-part of the hand-eye calibration matrix with ground truth values. They conducted only the eye-in-hand calibration evaluation experiments, and the input data format is exclusively 3D point clouds.

In our study, the imaging error is accounted for by conducting the static test, while the accuracy of our proposed method is evaluated through 3D reconstruction experiments. The comparison is presented in Table \ref{tab:comparison}, where \textit{None} indicates that it is not included in this work. The position and rotation errors along the X, Y, and Z axes are calculated based on the evaluation experiments conducted in their studies. These errors are defined with respect to the robot base frame and represent the potential offset from the actual calibration value. In addition, it is worth noting that the listed solutions only measure the algorithm's processing time and overlook the time spent on robot movement and human assistance, which can significantly increase the overall time required for hand-eye calibration. Our proposed method includes the entire process from the movement of the robot arm, which takes about 5 seconds, to 3D detection and extraction, point cloud registration, and final calibration calculation. As a result, the entire hand-eye calibration procedure can be completed within 6 seconds.

In addition, according to the results presented in Table \ref{tab:comparison}, the position and rotation error of our method is comparable to other methods. This difference could be due to their use of more precise sensors that produce less noise and fewer outliers. For example, a laser profile sensor Murali \cite{murali2021situ} with a Z-axis resolution of 0.092 to 0.488 mm and a Z-axis repeatability of $\pm \text{12} \mu m$. In general, line lasers and laser profile sensors offer higher accuracy than ToF cameras and structured light scanners. The device used in \cite{zhang2023novel} is a structured light camera with a repeatable accuracy of $\pm$ 0.15 mm. In \cite{xing2022reconstruction}, a high-end and high-accuracy Mechmind 3D sensor is employed to verify the accuracy of proposed method. However, the sensor used in our method has a depth repeatability of less than 3 mm and a point density of 0.9 mm. Therefore, our camera is the lowest quality device, producing more noise and outlier points compared to scanners typically used in other methods. However, it also demonstrates that our study has the potential to be easily extended and applied. Additionally, our method exhibits a faster runtime while maintaining comparable accuracy. Specifically, our method is twice as fast as the fastest among the other methods. The results indicate that our methodology demonstrates robustness and effectiveness in achieving accurate hand-eye calibration.



\begin{table*}[]
\centering
\caption{Comparison of other hand-eye calibration methods using 3D point clouds}
\label{tab:comparison}
\resizebox{\linewidth}{!}{%
\begin{tabular}{@{}lcccccc@{}}
\toprule
\multirow{2}{*}{Method \& Year} & \multicolumn{3}{c}{\multirow{2}{*}{Position error (mm)}} & \multirow{2}{*}{Rotation error (deg)} & \multirow{2}{*}{Runtime (s)} & \multirow{2}{*}{Camera type} \\
 & \multicolumn{3}{c}{} &  &  &  \\ \midrule
\multirow{2}{*}{Zhe \cite{wang2020efficient}, 2020} & (X:1.334 & Y:0.511 & Z:0.925) & \multirow{2}{*}{1.231} & \multirow{2}{*}{None} & \multirow{2}{*}{Line laser} \\
 & \multicolumn{3}{c}{1.108} &  &  &  \\
\multirow{2}{*}{Mingyang \cite{li2021robot}, 2021 } & (X:0.82 & Y:1.22 & Z:1.22) & \multirow{2}{*}{None} & \multirow{2}{*}{1227} & \multirow{2}{*}{Line laser} \\
 & \multicolumn{3}{c}{1.910} &  &  &  \\
\multirow{2}{*}{Murali \cite{murali2021situ}, 2021 } & (X:\textbf{0.701} & Y:\textbf{0.443} & Z:\textbf{0.366}) & \multirow{2}{*}{1} & \multirow{2}{*}{None} & \multirow{2}{*}{Laser profile} \\
 & \multicolumn{3}{c}{\textbf{0.906}} &  &  &  \\
Shiyu \cite{xing2022reconstruction}, 2022  & \multicolumn{3}{c}{1.7} & 0.4 & 10.9 & Structured light \\
Lei \cite{zhang2023novel}, 2023  & \multicolumn{3}{c}{1.84} & \textbf{0.325} & None & Structured light \\
Peters\cite{peters2024robot}, 2024  & \multicolumn{3}{c}{1.77} & 0.547 & 15984 & Structured light \\
\multirow{2}{*}{Ours} & (X:1.159 & Y:0.697 & Z:1.025) & 0.994 & \multirow{2}{*}{\textbf{\textless 1s + 5 (move)}} & \textbf{Low-cost Structured light} \\
 & \multicolumn{3}{c}{1.697} &  &  &  \\ \bottomrule
\end{tabular}%
}
\end{table*}

Additionally, it is worthy noting that several factors can influence the outcome of 3D reconstructions based on the hand-eye calibration. They are
\begin{enumerate}
    \item Errors in Robotic Movement: It can lead to inaccuracies in the pose of the robot flange, impacting both position and orientation accuracy.
    \item Incorrect Calibration: It disrupts the results of the calibrated point clouds captured in the camera frame.
    \item Noise Data: It leads to a distorted representation of the object's surface in the real world.
\end{enumerate}
In 1), it is essential to consider both the repeatability and accuracy of the robot during hand-eye calibration. The official documentation for the UR5e specifies a repeatability of $\pm$ 0.03 mm. Furthermore, accuracy, which assesses the discrepancy between desired and actual positioning, is critical. According to Ryo's work \cite{9981610}, the UR5e has a positioning accuracy of 0.1 mm, which might slightly affect the precision of the hand-eye calibration. Additionally, 3) is significant due to the depth accuracy of the 3D camera, which is up to 3 mm within our working distance, significantly impacting the calibration results. Therefore, the imaging error is considered in our study through the static test.

\subsection{Robotic Grasping}
\begin{figure*}[htbp]
    \centering
    \subfigure[40mm]{
        \includegraphics[width=0.2\linewidth]{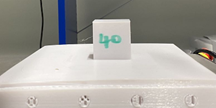}
        \label{fig:cube1}
    }
    \subfigure[30mm]{
        \includegraphics[width=0.2\linewidth]{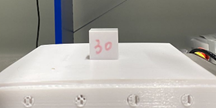}
        \label{fig:cube2}
    }
    \subfigure[25mm]{
        \includegraphics[width=0.2\linewidth]{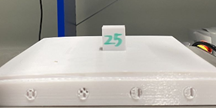}
        \label{fig:cube3}
    }
    \subfigure[20mm]{
        \includegraphics[width=0.2\linewidth]{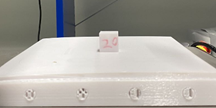}
        \label{fig:cube4}
    }
    \caption{A variety of printed cubes}
    \label{fig:grasping_objects}
    
    \subfigure[40 mm]{
        \includegraphics[width=0.2\linewidth]{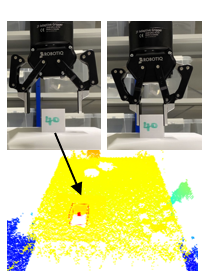}
    }
    \subfigure[30 mm]{
        \includegraphics[width=0.2\linewidth]{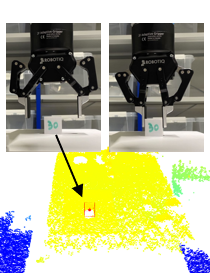}
    }
    \subfigure[25 mm]{
        \includegraphics[width=0.2\linewidth]{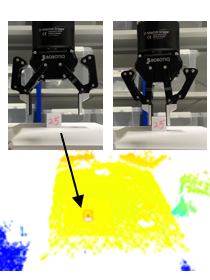}
    }
    \subfigure[20 mm]{
        \includegraphics[width=0.2\linewidth]{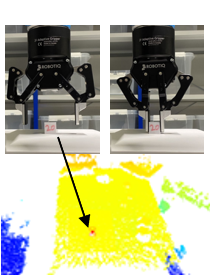}
    }
    \caption{Grasping positions for various size of cubes}
    \label{fig:grasping_position}
\end{figure*}

Robotic grasping is a common application involving interaction with objects, which requires hand-eye calibration. To further evaluate the performance of our hand-eye calibration, we designed a grasping experiment involving 3D printed cubes of various sizes, as illustrated in Fig. \ref{fig:grasping_objects}. The cubes vary in edge length from 20 mm to 40 mm. The hand-eye calibration in this robotic grasping system is implemented using our proposed method.

We begin by removing the points corresponding to the ground plane. The remaining points represent those of the cube. The grasping position is determined by calculating the center point of the 3D bounding box around these points, allowing for pick-up with a two-finger gripper (ROBOTIQ 2F-85). The estimated gripping positions are highlighted by red points, as shown in Fig. \ref{fig:grasping_position}. We conducted a total of 20 gripping trials for each cube size. The results indicate that these positions are accurately approached and successfully grasped by the gripper.

\subsection{3D Indoor Reconstruction}
\begin{figure*}[htbp]
    \centering
    \subfigure{
        \includegraphics[width=0.45\linewidth]{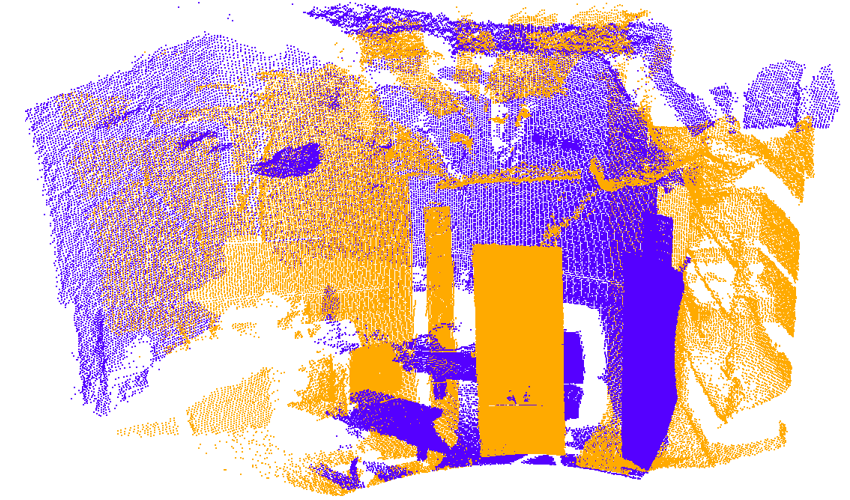}
    }
    \subfigure{
        \includegraphics[width=0.45\linewidth]{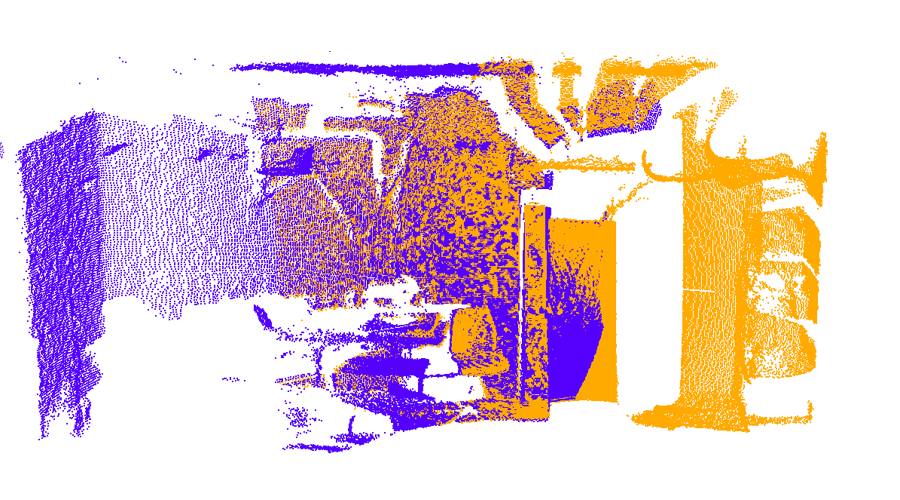}
    }\quad
    \subfigure{
        \includegraphics[width=0.45\linewidth]{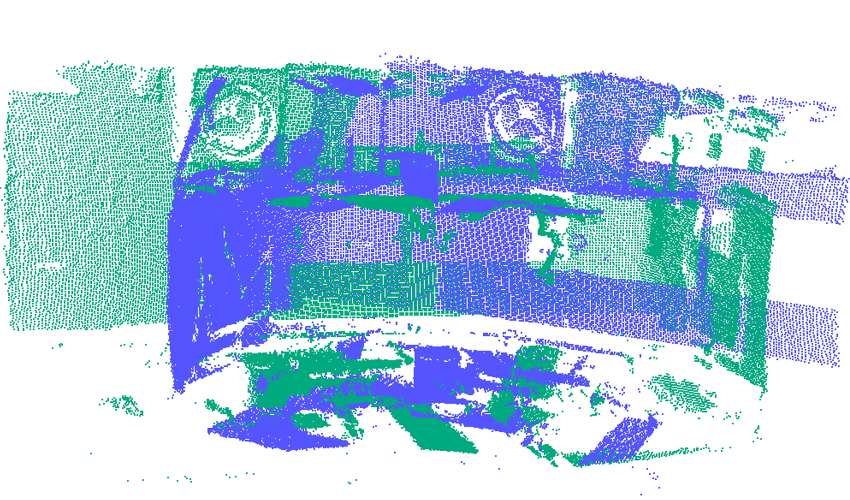}
    }
    \subfigure{
        \includegraphics[width=0.45\linewidth]{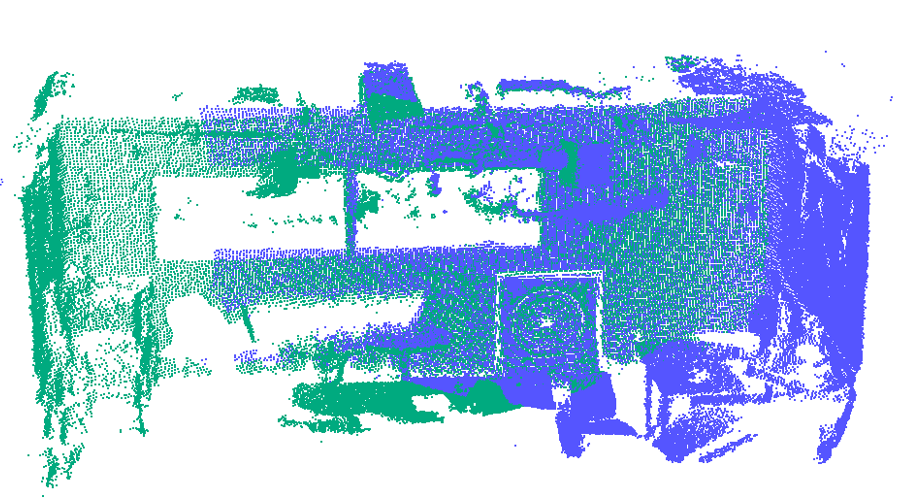}
    }\quad
    \subfigure{
        \includegraphics[width=0.45\linewidth]{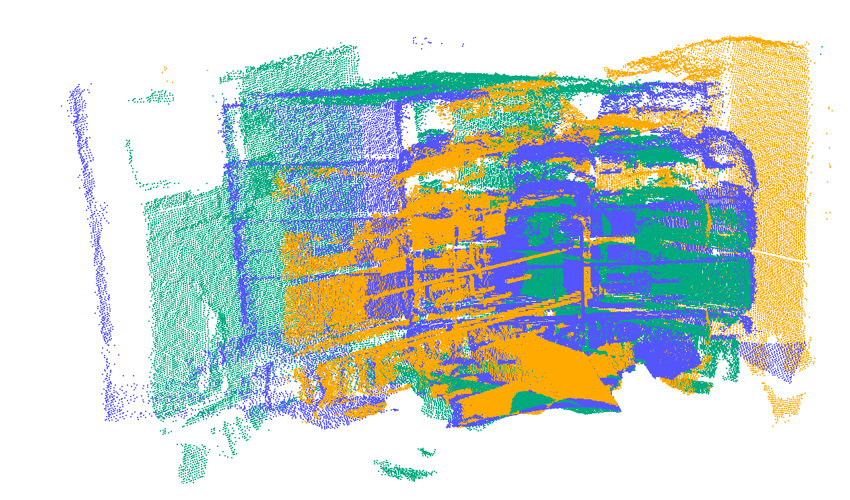}
    }
    \subfigure{
        \includegraphics[width=0.45\linewidth]{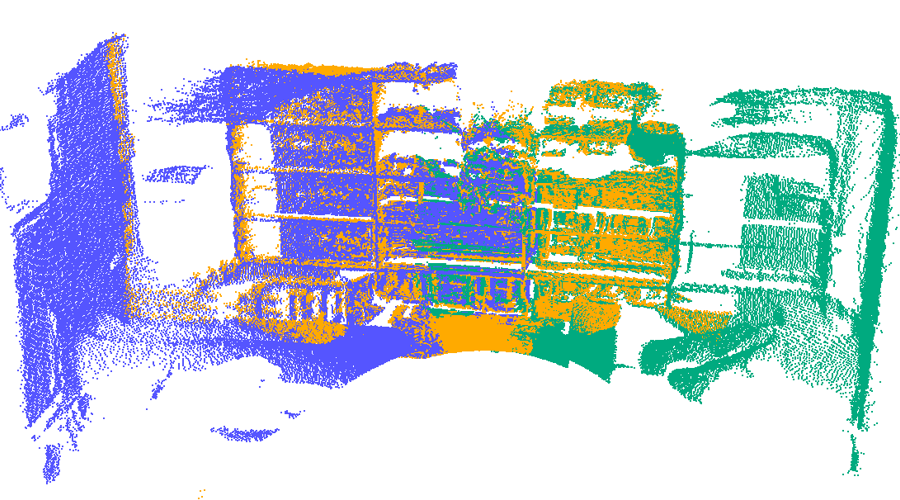}
    }\quad
    \subfigure{
        \includegraphics[width=0.45\linewidth]{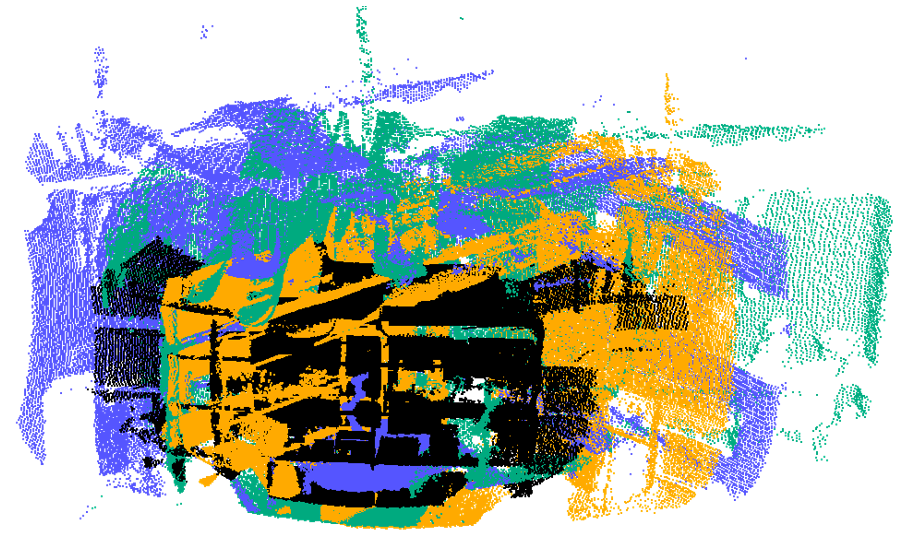}
    }
    \subfigure{
        \includegraphics[width=0.45\linewidth]{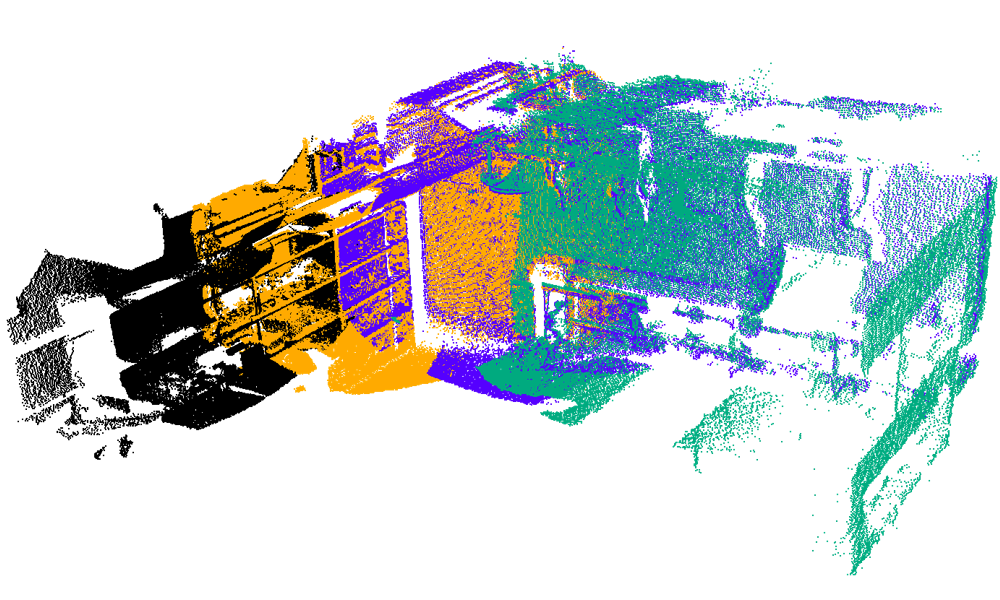}
    }
    \caption{Indoor 3D reconstruction: uncalibrated point clouds on the left and calibrated point clouds on the right. The calibrated point clouds around the robot can be aligned to form a comprehensive environment.}
    \label{fig.indoor}
\end{figure*}

 3D indoor reconstruction, which relies on hand-eye calibration, is widely used to enable robots to perceive and map their working environments. We conduct a 3D indoor reconstruction to demonstrate the utility of this technology. A 3D camera is mounted on the robot arm, and point clouds are acquired from two or more perspectives. Ideally, these calibrated point clouds can be aligned to construct a comprehensive environment around the robot. The results are depicted in Fig.\ref{fig.indoor} in which the left side of the figure displays the uncalibrated data, while the right side shows the calibrated point clouds. Each frame in the figure is highlighted in a different color.

\section{Conclusion} \label{conclusion}
This paper has proposed a novel approach to hand-eye calibration using only point clouds of the robot base. The principle behind this method is simple and effective. Specifically, it estimated a homogeneous transformation matrix between the sensor frame (3D camera) and the end-effector. This was achieved by learning-based framework that includes 3D detection and registration to estimate the 6D pose of the robot base. The 3D detection provided a bounding box by which the RoI, consisting of the points of the robot base, can be extracted. The purpose of the registration is to find the rigid relationship between the camera and the robot base frame by aligning the RoI with a reference model.

On the one hand, a learning-based detection framework, PV-RCNN++, has been employed to provide a prediction result with an average 86\% overlap with the real object. The robot base can be detected and extracted as an ROI within 0.09 seconds according to the experiments. On the other hand, a learning-based registration framework, PREDATOR (Point cloud REgistration with Deep Attention To the Overlap Region), has been employed to align the RoI with a 3D model of the robot base. This alignment, integrated with the ICP algorithm, can be achieved in half a second. Additionally, it outperforms other traditional registration methods in terms of RMSE, RRE, RTE, and OR according to the experiments. The robot base dataset used for training is generated in a virtual environment using hidden surface estimation, enabling its extension to other types of robot bases. In addition, a collaborative robot, UR5e, and a low-cost depth camera capable of producing point cloud data are utilized in our study. These devices are commonly found in SMEs and are representative of typical setups.

Furthermore, the robustness of our method was evaluated through three groups of experiments at six different joint configurations. The results demonstrate that our method can achieve expected performance at various configurations rather than relying on a strict and specific pose of the robot arm. Moreover, the accuracy of our method has been assessed through 3D reconstruction experiments, including static and dynamic tests. The static test was used to measure imaging errors. The results indicate position offsets of 1.159 mm, 0.697 mm, and 1.025 mm along the X, Y, and Z axes, and a rotation offset of 0.994 degrees. These findings are reliable based on the results obtained from the dynamic test with ICP. A robotic grasping and 3D indoor reconstruction demonstration were conducted at the end, which are common applications for collaborative robots in SMEs.

We compared our method with other 3D vision-based calibration methods using an experimental platform consisting of a collaborative robot and a low-cost camera. The camera used in our study has the lowest imaging accuracy compared to those used in other methods. Despite this, our experiments show that we achieved comparable performance at the fastest speeds. Our method can complete hand-eye calibration with a single robot arm movement, making it faster than traditional methods that typically require two or more movements. Additionally, the hand-eye calibration process runs quickly and can be implemented within 6 seconds, including robot pose adjustments. Although our method can also be benefited from the multiple movements, we make fast and reliable hand-eye calibration possible for collaborative robots.

Furthermore, our proposed calibration method is easily applicable to collaborative robots, given the accessible 3D model of the robot arm. For other industrial robots without open access to a 3D model, conducting a 3D reconstruction in advance to build the reference data for the robot base could be a potential solution.


In future studies, our proposed method will be evaluated for its suitability in various industrial applications, such as bin-picking, workpiece quality inspection, and robotic assembly. We believe that our approach could offer a fast and accurate solution for 3D vision-based hand-eye calibration. Additionally, exploring the potential use of our method with industrial robot arms like ABB and KUKA is an ongoing area of research. These robot arms have different shapes and materials for their robot bases.\\

\noindent \textbf{Acknowledgements} The support of Jungner Company and the Wenzhou Major Science and Technology Innovation Project (ZG2022011) is gratefully acknowledged. The authors would like to thank our teammate Yixiong Du for his wonderful collaboration and patient support.

\noindent \textbf{Conflicts of Interest} The authors declare that they have no known competing financial interests or personal relationships that could have appeared to influence the work reported in this paper.

\noindent \textbf{Code availability} The code and programming generated during and/or analyzed during the current study are available from the corresponding author upon reasonable request.

\noindent \textbf{Authors' Contributions} Leihui Li: Methodology, Implementation, Validation, Writing Original Draft. Xingyu Yang: Methodology, Review, and Editing. Riwei Wang and Xuping Zhang*: Supervision, Conceptualization, Methodology, Writing Original Draft.

\section*{Declarations}
\noindent \textbf{Ethical approval} This article does not contain any studies with human participants performed by any of the authors.
\noindent \textbf{Consent to Participate} Not applicable.
\noindent \textbf{Consent for Publication} Not applicable.

\bibliography{refs}
\end{document}